\documentclass[11pt]{article}
\usepackage[preprint]{acl}
\usepackage{times}
\usepackage{latexsym}
\usepackage{amsmath}
\usepackage[T1]{fontenc}
\usepackage[utf8]{inputenc}
\usepackage{microtype}
\usepackage{inconsolata}
\usepackage{graphicx}
\usepackage{booktabs}

\usepackage[utf8]{inputenc}
\usepackage{CJKutf8}
\usepackage{pifont}
\usepackage[table]{xcolor}
\usepackage{multirow}
\usepackage{subcaption}
\usepackage{amsfonts}
\usepackage{amssymb}
\usepackage{tabularx}
\usepackage{enumitem}
\usepackage[most]{tcolorbox}
\usepackage[dvipsnames]{xcolor}
\newcolumntype{C}{>{\centering\arraybackslash}X}

\title{Are LLMs Ready to Assist Physicians?\\
PhysAssistBench for Interactive Doctor-Patient-EHR Assistance}

\author{
  \bfseries\small
  Tianming Du\textsuperscript{1,2},
  Peijie Yu\textsuperscript{3},
  Sihan Shang\textsuperscript{1,2},
  Danli Shi\textsuperscript{4},
  My Linh Nguyen\textsuperscript{1,2}, \\
  \bfseries\small
  Shengbo Gao\textsuperscript{1,2},
  Guangyuan Li\textsuperscript{1,2},
  Yinghong Yu\textsuperscript{1,2},
  Yan Jiang\textsuperscript{1,5},
  Qianlong Zhao\textsuperscript{2,6}, \\
  \bfseries\small
  Behzad Bozorgtabar\textsuperscript{7},
  Shaoxiong Ji\textsuperscript{1,8},
  Jiazhen Pan\textsuperscript{9}, 
  Daniel Rueckert\textsuperscript{9,10},
  Jiancheng Yang\textsuperscript{1,2}\thanks{Corresponding author: Jiancheng Yang.},  \\
  \normalfont \small
  \textsuperscript{1}ELLIS Institute Finland, 
  \textsuperscript{2}Aalto University, 
  \textsuperscript{3}Tencent HY, 
  \textsuperscript{4}Hong Kong Polytechnic University, \\
  \small 
  \textsuperscript{5}University of Oulu, 
  \textsuperscript{6}Polytechnic University of Milan, 
  \textsuperscript{7}Aarhus University,
  \textsuperscript{8}University of Turku,\\
  \small
  \textsuperscript{9}Technical University of Munich,
  \textsuperscript{10}Imperial College London\\
  \small
  \texttt{\{du.tianming,jiancheng.yang\}@aalto.fi}
}

\begin{document}
\maketitle


\definecolor{dgreen}{RGB}{0,150,0}
\definecolor{orange}{RGB}{255,120,0}

\newif\ifdraft
\draftfalse

\ifdraft
  \newcommand{\JC}[1]{{\color{red}{\bf JC: #1}}}
  \newcommand{\jc}[1]{{\color{red} #1}}
  \newcommand{\NB}[1]{{\color{dgreen}{\bf NB: #1}}}
  \newcommand{\nb}[1]{{\color{dgreen} #1}}
  \newcommand{\TD}[1]{{\color{orange}{\bf TD: #1}}}
  \newcommand{\td}[1]{{\color{orange} #1}}
  \newcommand{\TODO}[1]{{\color{red}{\bf TODO: #1}}}
\else
  \newcommand{\JC}[1]{}
  \newcommand{\jc}[1]{ #1 }
  \newcommand{\NB}[1]{}
  \newcommand{\nb}[1]{ #1 }
  \newcommand{\TD}[1]{}
  \newcommand{\td}[1]{ #1 }
  \newcommand{\TODO}[1]{{\color{red}{\bf TODO: #1}}}
\fi

\newcommand{\bench}{\textsc{PhysAssistBench}}
\newcommand{\cmark}{\textcolor{green!60!black}{\ding{51}}}
\newcommand{\xmark}{\textcolor{red!70!black}{\ding{55}}}

\newcommand{\IL}{\textsc{il}}
\newcommand{\DG}{\textsc{dg}}
\newcommand{\CR}{\textsc{cr}}
\newcommand{\WU}{\textsc{wu}}

\newcommand{\NA}{\textsc{na}}
\newcommand{\PE}{\textsc{pe}}
\newcommand{\AAE}{\textsc{ae}}

\newcommand{\parag}[1]{\vspace{-2mm}\paragraph{#1}}
\newcommand{\ie}{\textit{i.e.}}
\newcommand{\eg}{\textit{e.g.}}
\newcommand{\etal}{\textit{et al.}}

\begin{abstract}

The most plausible near-term role of medical LLMs is to assist rather than replace physicians, yet current evaluations often test isolated capabilities: clinical knowledge, EHR system interaction, or patient communication. Physician assistance instead requires coordinating these capabilities within the same interaction, where physicians issue underspecified requests, patients describe symptoms ambiguously, and EHR systems demand precise tool use. We introduce \textbf{\bench{}}, a benchmark for interactive doctor-patient-EHR assistance. Built from real MIMIC-IV cases, \bench{} uses a scalable pipeline to construct \textit{agentic patients}: interactive, record-grounded agents that turn static EHR records into multi-turn clinical scenarios while preserving clinical factuality. \bench{} provides a curated bilingual evaluation set of 1,296 manually reviewed and physician-validated turns. Experiments with leading LLMs show that current models remain unreliable in this setting, which exposes a key bottleneck for clinical LLMs: reliable assistance requires coordination across knowledge, communication, and systems, not isolated gains in any of them.
\end{abstract}


\section{Introduction}
\label{sec:intro}



\begin{figure*}[t]
\centering
\includegraphics[width=0.95\textwidth]{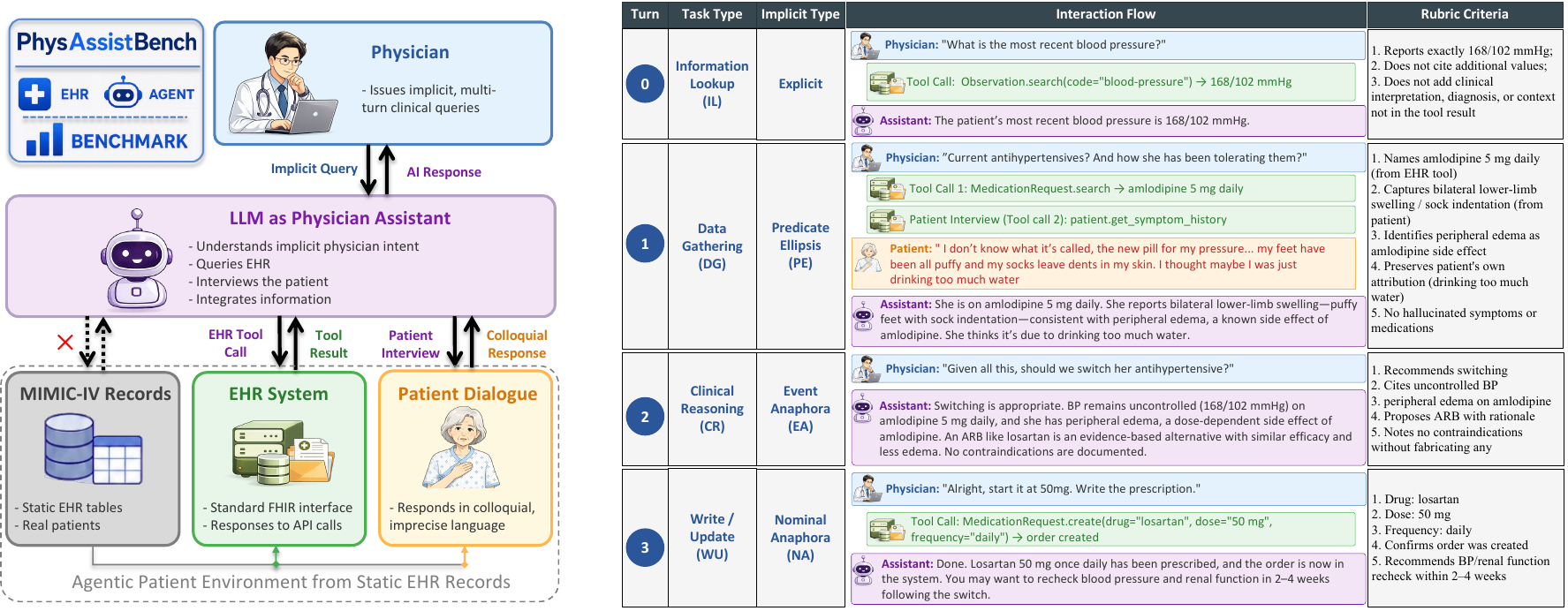 }
\caption{
\textbf{Left:} \bench{} evaluates LLMs as physician assistants, not physicians: the assistant follows physician requests while interacting with a record-grounded FHIR-based EHR system and a dialogue patient.
A multi-agent pipeline transforms static MIMIC-IV records into this \textit{agentic patient} environment, exposing them through standardized FHIR interfaces rather than direct record access.
\textbf{Right:} A representative hypertension case in \bench{}.
The 4-turn session progresses from explicit lookup to implicit assistance involving patient dialogue, clinical reasoning, and EHR actions, with each turn paired with its interaction flow and rubric criteria.
}
\label{fig:session-example}
\vspace{-10px}
\end{figure*}

LLMs have generated substantial optimism for clinical AI
\cite{thirunavukarasu2023llm,moor2023foundation,rajpurkar2022ai}.
Much of this optimism comes from their strong performance on medical examinations and question-answering benchmarks
\cite{kung2023chatgpt,singhal2025medpalm2}, motivating visions of LLMs as a new ``front door to healthcare''
\cite{nhsengland2024frontdoor,kyle2025ai}.
However, recent studies suggest that such performance may not transfer to interactive clinical use.
\citet{laban2025lost} found that shifting from fully specified single-turn prompts to multi-turn, under-specified interactions caused an average 39\% performance drop.
Similarly, \citet{bean2026reliability} found that LLMs performed strongly when tested alone, but failed to improve user performance in a randomized medical self-assessment study, identifying user interaction as a key barrier.

These findings reveal a gap in current evaluation practice: practical failures may arise not only from insufficient medical knowledge, but from failures of interaction.
As summarized in Table~\ref{tab:benchmark_comparison}, prior medical LLM benchmarks mostly evaluate three isolated roles: \textit{knowledge}, where models act as medical experts; \textit{system}, where models retrieve or manipulate clinical records; and \textit{communication}, where models interact with patients or generate clinical text.
These roles are useful, but they miss the most plausible near-term deployment setting: assisting physicians under human oversight, as emphasized by ethical and regulatory expectations~\cite{who2021ethics,fda2025ai}.
Technically, this setting is also where recent studies expose a key bottleneck: interaction~\cite{laban2025lost,bean2026reliability}.
Physician assistance is not static question answering, but interactive coordination across incomplete physician intent, ambiguous patient information, and precise EHR actions.

Figure~\ref{fig:session-example} illustrates the setting studied in this paper.
Even for medical professionals, physician requests are often context-dependent, elliptical, and spread across turns.
The assistant must map these implicit requests to two distinct interfaces: EHR systems requiring precise tool calls, and patients providing colloquial, incomplete, and clinically imprecise information.
It must decide when to query the EHR, when to ask the patient, and how to integrate evidence into a physician-facing response.
As will be discussed in Section~\ref{sec:related}, this combination is not captured by existing benchmarks.


We introduce \bench{}, a benchmark for interactive doctor-patient-EHR assistance.
Evaluating such interactions requires more than static clinical records \cite{li2025srrgbench}: it must provide patients who can respond across turns, EHR systems that can be queried through structured tools, and physician requests that evolve with context.
To make this scalable while preserving clinical grounding, \bench{} repurposes real MIMIC-IV~\cite{johnson2023mimic} cases through a multi-agent synthetic data pipeline.
The pipeline plans clinically plausible scenarios from eligible records and constructs \textit{agentic patients}: interactive, record-grounded agents that turn static EHR cases into multi-turn clinical scenarios.
Unlike unconstrained patient simulation, these agents are grounded in existing EHR evidence; unsupported cases are filtered rather than counterfactually rewritten.


The formulated benchmark contains 324 multi-turn sessions and 1,296 turns.
Eight trained annotators collectively verified and corrected the annotated fields of every turn.
Separately, a board-certified physician and the trained annotators assessed query realism and trajectory coherence for all 324 sessions. 
It spans 4 clinical scenarios (Diagnostic Workup, Med Safety, Treatment Response and Discharge Planning), 4 tasks (Information Lookup, Data Gathering, Clinical Reasoning, and Write/Update), and 3 physician-query implicitness subtypes: Nominal Anaphora (\NA{}), Predicate Ellipsis (\PE{}), and Abstract Event Anaphora (\AAE{}).
Each session uses a standardized FHIR R4 tool set and is evaluated in English and Chinese.
Turn-level rubrics are provided for stable and interpretable assessment.
Figure~\ref{fig:session-example} shows a representative 4-turn session, illustrating how implicit physician requests, patient ambiguity, and EHR precision co-occur within the same workflow.


Our contributions are threefold.
First, we formulate interactive physician assistance as a coordination problem across clinical knowledge, patient communication, and EHR systems, a setting overlooked by existing evaluations of isolated roles.
Second, we develop a scalable multi-agent synthetic data pipeline that repurposes static MIMIC-IV records into \textit{agentic patients}, enabling clinically grounded multi-turn doctor-patient-EHR scenarios.
Third, we release \bench{}, a manually reviewed and physician-validated benchmark of 324 sessions and 1,296 turns, and show that leading LLMs are not yet reliable physician assistants, especially when resolving physician intent, handling patient ambiguity, issuing grounded EHR queries, and integrating evidence across sources.
Dataset and code are made publicly available\footnote{\url{https://github.com/HINTLab/PhysAssistBench}}.



\begin{table*}
  \centering
  \scriptsize
  \renewcommand{\arraystretch}{1.15}
  \begin{tabular*}{0.90\textwidth}{@{\extracolsep{\fill}} l c c c c c@{}}
  \toprule
    \textbf{Benchmark}
& \textbf{Evaluation Focus}
& \textbf{Implicit Queries}
& \textbf{Patient Interaction}
& \textbf{Tool Use}
& \textbf{Turns}\\
  \midrule
   PubMedQA~\cite{jin2019pubmedqa} & Knowledge & \xmark & \xmark & \xmark & Single \\
   EHRSQL~\cite{lee2022ehrsql}                   & System                   & \xmark & \xmark & \xmark & Single \\
   EHRXQA~\cite{bae2023ehrxqa}                   & System, Knowledge        & \xmark & \xmark & \xmark & Single \\
   EHRAgent~\cite{shi-etal-2024-ehragent}        & Knowledge                & \xmark & \xmark & \cmark & Single \\
   EHRNoteQA~\cite{kweon2024ehrnoteqa}           & Knowledge                & \xmark & \xmark & \xmark & Single \\
   ClinicalBench~\cite{chen2024clinicalbench}    & Knowledge                & \xmark & \xmark & \xmark & Multi  \\
   ER-REASON~\cite{mehandru2025er}               & Knowledge                & \xmark & \xmark & \xmark & Single \\
   CSEDB~\cite{wang2025csedb}                    & Knowledge                & \xmark & \xmark & \xmark & Single \\
   MedThink-Bench~\cite{zhou2025automating}      & Knowledge                & \xmark & \xmark & \xmark & Single\\
   MedAgentsBench~\cite{tang2025medagentsbench}  & Knowledge                & \xmark & \xmark & \cmark & Single \\
   MedAgentBench~\cite{jiang2025medagentbench}   & System                   & \xmark & \xmark & \cmark (FHIR)& Single \\
   FHIR-AgentBench~\cite{lee2025fhiragentbench}  & System                   & \xmark & \xmark & \cmark (FHIR) & Single \\
   AgentClinic~\cite{schmidgall2026agentclinic}  & Knowledge, Communication & \xmark & \cmark & \cmark & Multi \\
   MedAgentGym~\cite{xu2026medagentgym}          & System, Knowledge        & \xmark & \xmark & \cmark  & Multi \\

  \midrule
   \textbf{\bench{}} (Ours) & \textbf{Assistance}
     & {\cmark} & {\cmark} & \cmark (FHIR) & Multi \\
  \bottomrule
  \end{tabular*}
\caption{
Comparison of existing EHR and medical agent benchmarks.
\textit{Evaluation Focus} denotes the primary capability tested:
\emph{knowledge} for medical expertise,
\emph{system} for EHR or tool use, and
\emph{communication} for patient dialogue.
\bench{} is the only benchmark with integrated \emph{assistance} focus across all three interaction dimensions.
\textit{Implicit Queries}, \textit{Patient Interaction}, \textit{Tool Use}, and \textit{Turns} denote underspecified physician requests, simulated-patient interaction, executable tools with FHIR noted when applicable, and single- or multi-turn evaluation.
}
  \label{tab:benchmark_comparison}

\vspace{-10px}
  
  \end{table*}

\section{Related Work}
\label{sec:related}

\subsection{EHR and Medical Agent Benchmarks}



 Table~\ref{tab:benchmark_comparison} compares existing  benchmarks by evaluation focus and interaction dimensions.
Most prior benchmarks evaluate isolated capabilities rather than integrated physician-assistant workflows.

A large body of work focuses on \textit{knowledge}, testing medical reasoning over questions, notes, records, or agent-style clinical tasks~\cite{jin2019pubmedqa,shi-etal-2024-ehragent,kweon2024ehrnoteqa,chen2024clinicalbench,mehandru2025er,wang2025csedb,zhou2025automating,tang2025medagentsbench}.
These benchmarks assess clinical expertise, but mostly assume fully specified, static inputs without patient or EHR interaction across turns.
Another line focuses on \textit{system} interaction, evaluating EHR retrieval or clinical tool use~\cite{lee2022ehrsql,bae2023ehrxqa,jiang2025medagentbench,lee2025fhiragentbench,xu2026medagentgym}.
However, they typically treat EHR access as standalone tool use and do not test implicit references, argument elision, or cross-turn carry-over.
A third line focuses on \textit{communication}, where models interact with patients or generate clinical text.
AgentClinic~\cite{schmidgall2026agentclinic} supports multi-turn doctor-patient interaction and is closest to the dialogue side of our setting, but does not jointly evaluate structured EHR tool use, patient dialogue, and implicit physician requests.

Overall, existing benchmarks cover important pieces of clinical LLM evaluation, but not the integrated \textit{assistance} setting, where an LLM must coordinate \textit{knowledge}, \textit{communication}, and \textit{system} interaction under evolving physician instructions. \citet{luo2026clinical} also articulate a research vision 
for evaluating clinical LLMs through realistic EHR interfaces 
in dynamic, interactive clinical settings, sharing our critique 
of current evaluation paradigms. 
As shown in Table~\ref{tab:benchmark_comparison}, \bench{} is the only benchmark covering implicit queries, patient interaction, FHIR-based EHR tool use, and multi-turn evaluation together.
This coverage is made possible by a scalable multi-agent synthetic data pipeline that turns MIMIC-IV records into interactive doctor-patient-EHR scenarios.

\subsection{Medical Synthetic Data}

Medical synthetic data has become an important strategy for addressing data scarcity, privacy, and bias in healthcare.
Most prior work follows a \textit{generative} synthesis paradigm, creating realistic medical images, clinical text, tabular EHR data, or multimodal patient records for training and evaluation via generative models~\cite{fansi2022ddxplus,li2023synthetic,seo2024diagesc,zhang2025lefusion,zhang2026multimodal}.
Beyond sample generation, recent work also studies \textit{agentic} data synthesis, where LLMs or multi-agent systems construct clinical dialogues through prompt-encoded constraints~\cite{das2024synthetic} or clinician-patient role simulation grounded in EHR-derived evidence~\cite{wang2024notechat,almutairi2024synthetic,xu2026rcbsf}.

\bench{} differs by using synthetic data to construct an interactive evaluation environment.
Its scalable multi-agent synthetic data pipeline plans clinically plausible scenarios from real records and constructs \textit{agentic patients}, which are grounded in existing EHR evidence rather than unconstrained simulation; unsupported cases are filtered instead of counterfactually rewritten.
The resulting evaluation set is further  manually reviewed and physician-validated for reliability.

\subsection{Tool Use and Function Calling}


General-domain tool-use benchmarks evaluate API selection, function calling, and tool-mediated task completion~\cite{li2023api,qin2023toolllm,krishna2025fact,yu2026benchmarking}.
While useful for studying agent tool use, they mostly assume general-purpose domains and do not model clinical tool schemas, FHIR-based EHR actions, or implicit physician queries.
\bench{} instead embeds tool use in an interactive doctor-patient-EHR workflow requiring coordination between FHIR-based EHR tools, ambiguous patient dialogue, and evolving physician instructions.

\subsection{Ellipsis and Anaphora in Dialogue}


Ellipsis and anaphora are pervasive in natural dialogue and have long been studied in linguistics and NLP~\cite{gerber2010beyond,lee2017end,lee2018higher,marasovic2017mention,kolhatkar2014resolving}.
They cover phenomena such as omitted predicates or arguments, entity references, and abstract event references, all of which require context-dependent interpretation.
In clinical communication, especially physician instructions, such implicitness is common: clinicians often rely on shared context, abbreviate repeated actions, and refer back to prior findings across turns.
Despite this, existing EHR and medical agent benchmarks typically formulate queries as explicit, standalone instructions, leaving implicit physician requests overlooked.

\section{\bench}
\label{sec:benchmark}

\subsection{Overview}

\bench{} is a multi-turn benchmark built on real MIMIC-IV patient data
that evaluates LLMs on realistic doctor-patient-EHR interaction.
It comprises 324 sessions and 1,296 turns, evaluated in both English
and Chinese using a standardized FHIR R4 tool set~\cite{mandel2016smart}.

\subsection{Benchmark Structure}
\label{sec:structure}

\paragraph{Sessions and Turns.}
The primary unit of evaluation is a \emph{session}: four consecutive
turns (Turn~0--Turn~3) simulating a single clinical encounter.
Turns are not independent: each builds on prior retrieved data and
established context, requiring the assistant to track what has already
been answered and reason across the full conversation.

\paragraph{Task Types.}
Each turn is labeled among four task types reflecting the
clinical intent of the physician's query.
\textit{Information Lookup (\IL{})} retrieves a single clinical fact
via one EHR read tool.
\textit{Data Gathering (\DG{})} collects information from two or more
sources (EHR tools, patient-interview tools) with calls
that may be parallel or adaptive.
\textit{Clinical Reasoning (\CR{})} combines retrieved findings with
medical knowledge to produce a clinical recommendation optionally involving tool calls.
\textit{Write/Update (\WU{})} executes a FHIR write operation whose
parameters are inferred from session context.
Turn~0 is always \IL{}, anchoring the session in a concrete data
request; Turn~3 always ends in \DG{}, \CR{}, or \WU{}.

\paragraph{Physician-Query Implicitness Types.}
Task type captures \emph{what} is being requested; the implicitness
type captures \emph{how} the physician expresses it.
As shared context accumulates, physicians naturally abbreviate:
referring back to prior entities by pronoun, dropping predicates, or
compressing the clinical picture into a brief phrase.
\bench{} encodes three implicitness types assigned to Turns~1--3.
\textit{Nominal Anaphora (\NA{}):} a pronoun or noun phrase refers to
a specific entity from a prior turn.
\textit{Predicate Ellipsis (\PE{}):} the verb and arguments are
omitted; only the new focus item is stated.
\textit{Abstract Event Anaphora (\AAE{}):} a phrase such as ``given
all this'' refers to the accumulated clinical picture.
Examples are in Table~\ref{tab:subtypes}.

\paragraph{Tool Set.}
The assistant draws on two tool families. \textit{EHR tools} are FHIR R4 read and write operations over
MIMIC-IV resources.
\textit{Patient-interview tools} elicit subjective information absent
from structured records (chief complaint, symptom history, medication
adherence, and functional status) with responses pre-generated from
the same admission records as the EHR snapshot and held fixed
at evaluation time.
Treating patient interaction as a tool call unifies EHR queries and patient questions under a single decision interface and enables deterministic evaluation via response replay. The full tool inventory is provided in Appendix~\ref{app:tools}.

\paragraph{Clinical Scenarios.}
The task and implicitness types instantiate across four
clinical scenarios spanning the physician's core decision cycle:
Diagnostic workup, Medication safety review (Med Safety), treatment response
monitoring (Treatment Response), and discharge planning.
Each scenario constrains which FHIR resources and tool
combinations are exercised, ensuring structural distinctiveness.
The scenario design is inspired by~\citet{jiang2025medagentbench,lee2025fhiragentbench,liu2026benchmarking}.

\paragraph{Data-Richness Tiers.}
Each session is assigned among three \emph{data-richness tiers}
reflecting the depth of structured evidence in the
MIMIC-IV record: a coverage dimension rather than a difficulty
ranking.
\emph{Data-sparse} sessions require a single lab value or
prescription.
\emph{Data-moderate} sessions require a trend-bearing time series or
a co-occurring drug-lab pair.
\emph{Data-rich} sessions require multi-dimensional evidence such as
multiple drug-monitoring pairs or multi-system differential findings.
Tier eligibility is verified offline against raw MIMIC-IV CSV files
without LLM involvement (Appendix~\ref{app:pipeline}).
Complete scenario-by-tier, task-by-turn, and implicitness-by-turn
distributions are reported in Appendix~\ref{sec:full-stats},
Tables~\ref{tab:dist-scenario-tier}--\ref{tab:dist-subtype}.


\begin{figure*}[!htbp]
    \centering
    \includegraphics[width=0.95\linewidth]{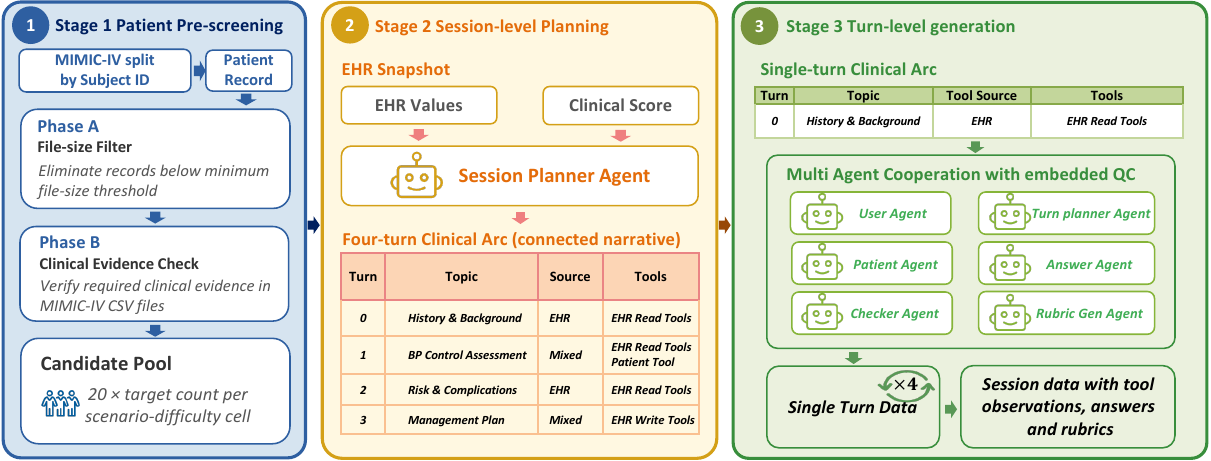}
    \caption{Multi-agent data synthesis pipeline. A static MIMIC-IV record is transformed into a grounded,
        multi-turn benchmark entry in the \textit{agentic patient} environment, through patient pre-filtering, session-level planning, and turn-level generation with embedded quality control.
    }
    \label{fig:agentic_pipeline}
    \vspace{-10px}
\end{figure*}

\subsection{Quality Assurance}
\label{sec:quality-assurance}

All 324 sessions underwent physician validation for clinical plausibility,
query realism, and trajectory coherence. The session-level realism assessment
yielded an overall mean of 4.37/5 and a minimum mean of 3.5, with no session
falling below the quality threshold. Separately, eight trained
annotators verified and corrected all turn-level annotated fields, including
the implicitness labels, rubric items, and gold tool calls, across all 1,296
turns. Appendix~\ref{sec:human-eval} documents the reviewer qualifications,
review and correction protocol, and complete annotation results.
Appendix~\ref{sec:realism-assessment} and
Tables~\ref{tab:realism-by-scenario}--\ref{tab:lowest-realism-session}
report the full-benchmark realism analysis.

\subsection{Evaluation}
\label{sec:eval}

\paragraph{Task Formalization.}
We formulate each \bench{} session
as a finite-horizon partially observable Markov decision process
(POMDP) $(\mathcal{S},\mathcal{A},\mathcal{O},T,R,\mathcal{U})$, with
state space $\mathcal{S}$, action space $\mathcal{A}$, observation
space $\mathcal{O}$, transition function
$T:\mathcal{S}\times\mathcal{A}\to\mathcal{S}\times\mathcal{O}$,
reward $R:\mathcal{S}\to[0,1]$, and physician-instruction space
$\mathcal{U}$.
The state decomposes as
$\mathcal{S}=\mathcal{S}_{\text{ehr}}\cup
\mathcal{S}_{\text{pat}}\cup\mathcal{S}_{\text{ctx}}$:
$\mathcal{S}_\text{ehr}$ is the MIMIC-IV patient snapshot frozen at $t_\text{anchor}$ and hidden from the agent; $\mathcal{S}_\text{pat}$ is the patient-interview state, constructed from the same admission-period records as $\mathcal{S}_\text{ehr}$ and fixed at pipeline initialization time.
$\mathcal{S}_{\text{ctx}}$ is the accumulated dialogue history
$H_t=((q_0,y_0),\ldots,(q_{t-1},y_{t-1}))$.
The action space partitions as
$\mathcal{A}=\mathcal{A}_{\text{ehr}}\cup
\mathcal{A}_{\text{pat}}\cup\mathcal{A}_{\text{nl}}$: FHIR R4 tool
calls (read and write), patient-interview tool calls, and the
natural-language answer $y_t$ returned to the physician.
The instruction space $\mathcal{U}$ contains the physician's queries
$q_t$.
A session is the tuple
$\sigma=(t_{\text{anchor}},q_0,q_1,q_2,q_3)$ paired with gold tool
calls and gold answers; at each turn the agent produces
$(\tau_t,y_t)$, where $\tau_t$ is the executed tool-call trace and
$y_t$ is the response.

\paragraph{Rubric-Based Scoring.}
Each turn is scored by an LLM judge on several independent rubric items
designed for the turn's task type.
Items cover factual accuracy of reported values, correctness of
clinical interpretation, appropriate integration of prior session
context, and absence of hallucinated facts.
The score for turn $t$ is the fraction of items passed,
$r_t\in[0,1]$; the session-level Rubric Score is
$\bar r(\sigma)=\tfrac{1}{4}\sum_{t=0}^{3}r_t$ and the corpus-level
\textit{mean rubric score (mRS)} reported in Table~\ref{tab:main_results} is
$\mathbb{E}_{\sigma}[\bar r(\sigma)]$.

\paragraph{Pass@Turn and Pass@Session.}
The continuous rubric score is binarised at threshold $\tau$ to
yield 2 reliability metrics.
\textit{Pass@Turn} (\textit{@T}$_\tau$) is the fraction of turns whose
rubric score meets the threshold:
\[
\text{@T}_\tau
=
\frac{
\sum_{\sigma}\sum_{t=0}^{3}
\mathbf{1}[r_t \ge \tau]
}{
4|\mathcal{D}|
}.
\]
\textit{Pass@Session} (\textit{@S}$_\tau$) is the fraction of sessions
in which all four turns meet the threshold:
\[
\text{@S}_\tau
=
\frac{
\sum_{\sigma \in \mathcal{D}}
\mathbf{1}\!\left[
\min_{t\in\{0,1,2,3\}} r_t \ge \tau
\right]
}{
|\mathcal{D}|
}.
\]
We report $\tau\in\{0.60, 0.75\}$. $|\mathcal{D}|$ is the number of sessions.
@S is the demanding metric: single weak turn fails entire
session, reflecting the multi-turn, context-dependent
character of clinical practice.

\paragraph{Tool Use Evaluation}
Tool invocation is evaluated implicitly through rubric-based scoring: a response grounded in correct EHR values necessarily required correct tool calls to retrieve them. This also accommodates cases where a model deviates from the gold tool trajectory but still produces a clinically correct answer, as the rubric rewards correctness rather than procedural conformity.

\section{Agentic Patient Environment from Static EHR Records}
\label{sec:agentic_pipeline}


We design the pipeline around a generative principle: each patient's EHR record, rather than serving as a static lookup table, acts as the \emph{driving participant} of an \emph{agentic patient} environment.
Given one admission's data, a coordinated set of agents decides what clinical questions the record can support, plans the required EHR and patient-tool interactions, executes real FHIR queries against the actual data, and produces verifiable gold responses without human authorship of the scenario itself.
The same record can therefore generate different conversations across scenarios, difficulty tiers, and languages, making the pipeline scalable while remaining grounded in which specific admission contains.
In this sense, the pipeline turns static EHR records into interactive \emph{agentic patients} capable of driving complex doctor-patient-EHR interactions.

As illustrated in Figure~\ref{fig:agentic_pipeline}, the pipeline comprises
3 stages.
\textit{(1) Patient pre-filtering} applies a two-stage offline filter:
first file-size thresholds, then scenario- and tier-specific content
checks ensuring every retained patient has sufficient EHR evidence
before any LLM is invoked.
\textit{(2) Session planning} reads the patient's EHR snapshot and
produces a coherent four-turn clinical arc specifying the per-turn
topic, FHIR tools or patient tools, transforming isolated tabular 
data into a connected clinical narrative.
\textit{(3) Turn-level generation} instantiates each planned turn
through multi-agent cooperation with tool-call planning, FHIR execution against real
data, and gold answer generation, with three embedded quality gates
rejecting hallucinated, structurally invalid, or clinically unsafe
outputs.

The pipeline scales by instantiating multiple sessions from a single patient 
record across scenarios, difficulty tiers, and languages; quality is enforced 
by a three-stage checker that filters hallucinations, structural errors, and 
unsafe content. Full pipeline details (Figure~\ref{fig:data_generation_pipeline}), prompt templates, and per-gate failure 
statistics are provided in Appendix~\ref{app:pipeline}.



\section{Experiments}
\label{sec:experiments}

\subsection{Models}
We benchmark 5 closed-source models: GPT-5.4, GPT-5.4 mini \cite{openai2026gpt54thinking}, Claude-Opus-4.7 \cite{claudeopus47_2026}, Gemini-3.1-Pro \cite{gemini31pro_blog_2026}, Seed-1.8\cite{seed18_2026} and 9 open-weight models: DeepSeek-V4-Pro and DeepSeek-V4-Flash \cite{deepseekv4_2026}, Qwen3.5 series models \cite{qwen35_2026},  GLM-5 \cite{glm5_2026}, Kimi-K2.6 \cite{kimik26_2026}, MiniMax \cite{chen2026minimax}. For each model we run the full English and Chinese benchmark
(324 sessions $\times$ 4 turns $\times$ 2 languages = 2{,}592 turns), under identical evaluation conditions.

\paragraph{Eval Configuration \& Judge Model.}
All models run with reasoning (``thinking'') mode enabled
(\texttt{reasoning\_effort=high} for the GPT-5 series),
$\text{temperature}=0.2$, and a maximum of 16 tool calls per turn,
with all 18 EHR, patient-interview and control tools available at every turn.
Rubric scoring is performed by a fixed GPT-5.4-mini judge which is identical
across all models. It receives the model's answer, EHR ground-truth, and rubric items, returning binary scores with reasoning for each item.

\paragraph{Judge Reliability.}
Judge reliability was measured on GLM-5 outputs from 108 sessions per language, covering 3,864 rubric items. The fixed GPT-5.4-mini judge achieved 95.37\% agreement with trained-annotator labels (95\% CI: $\pm$0.66 percentage points), with agreement exceeding 91\% across every reported task-, language-, and rubric-category subset. Appendix~\ref{sec:judge-reliability} and Tables~\ref{tab:judge-agreement-task}--\ref{tab:judge-agreement-category} provide the protocol and breakdowns.

\subsection{Main Results}
\begin{table*}[!htbp]
\centering\footnotesize
\setlength{\tabcolsep}{4.5pt}
\renewcommand{\arraystretch}{0.95}
\begin{tabular}{l ccccc ccccc@{}}
\toprule
\multirow{3}{*}{\textbf{Model}}
  & \multicolumn{5}{c}{\textbf{English}}
  & \multicolumn{5}{c}{\textbf{Chinese}} \\
\cmidrule(lr){2-6} \cmidrule(l){7-11}
  & \multirow{2}{*}{mRS}
  & \multicolumn{2}{c}{Pass@Turn}
  & \multicolumn{2}{c}{Pass@Session}
  & \multirow{2}{*}{mRS}
  & \multicolumn{2}{c}{Pass@Turn}
  & \multicolumn{2}{c}{Pass@Session} \\
\cmidrule(lr){3-4}\cmidrule(lr){5-6}
\cmidrule(lr){8-9}\cmidrule(l){10-11}
  & & $\tau{=}.60$ & $\tau{=}.75$ & $\tau{=}.60$ & $\tau{=}.75$
  & & $\tau{=}.60$ & $\tau{=}.75$ & $\tau{=}.60$ & $\tau{=}.75$ \\
\midrule
\multicolumn{11}{@{}l}{\textbf{Closed-Source Model} \scriptsize\cite{claudeopus47_2026,openai2026gpt54thinking,gemini31pro_blog_2026, seed18_2026}} \\ 

\midrule
Claude-Opus-4.7  & \color{blue}{68.3} & \color{red}{67.7} & 52.5 & \color{red}{23.5} & \color{red}{8.0} & \color{blue}{69.9} & \color{blue}{70.1} & \color{blue}{55.1} & \color{red}{26.9} & \color{red}{9.0} \\
GPT-5.4-high         & 64.1 & 63.4 & 42.3 & 17.3 & 2.2 & 65.9 & 63.0 & 43.8 & 18.8 & 4.9 \\
Gemini-3.1-Pro       & 62.1 & 59.9 & 40.3 & 11.1 & 1.5 & 68.0 & 68.8 & 48.5 & 22.5 & 4.6 \\
Seed-1.8             & 67.4 & 66.4 & 46.5 & 18.5 & 3.1 & 68.7 & 67.4 & 48.0 & 20.4 & 6.8 \\
GPT-5.4-mini                            & 57.8 & 53.8 & 37.6 & 8.0  & 1.2 & 61.5 & 58.7 & 41.7 & 11.7 & 2.2 \\
\midrule
\multicolumn{11}{@{}l}{\textbf{Open-Weight Model} \scriptsize\cite{glm5_2026, kimik26_2026, deepseekv4_2026, chen2026minimax, qwen35_2026}} \\
\midrule
GLM-5                  & \color{red}{69.4} & \color{blue}{67.6} & \color{red}{53.1} & \color{blue}{21.3} & \color{blue}{6.5} & \color{red}{71.5} & \color{red}{71.3} & \color{red}{55.3} & \color{blue}{26.5} & \color{blue}{8.3} \\
Kimi-K2.6             & 68.0 & 66.6 & 51.7 & 15.7 & 4.6 & \color{blue}{69.9} & 69.7 & 53.4 & 24.7 & \color{blue}{8.3} \\
DS-V4-Pro         & 68.1 & 66.7 & \color{blue}{52.9} & 20.7 & \color{blue}{6.5} & 64.8 & 63.4 & 48.8 & 19.4 & 5.6 \\
DS-V4-Flash       & 64.3 & 62.7 & 49.1 & 17.0 & 4.6 & 60.6 & 59.7 & 46.2 & 19.4 & 7.1 \\

MiniMax-M2.7      & 63.2 & 60.4 & 43.4 & 10.8 & 2.2 & 66.8 & 63.9 & 47.3 & 13.3 & 4.3 \\
Qwen3.5-35B-A3B     & 66.3 & 65.0 & 49.4 & 18.8 & 3.7 & 65.8 & 63.3 & 50.0 & 17.3 & 5.9 \\
Qwen3.5-27B                            & 58.3 & 55.6 & 40.2 & 11.1 & 1.2 & 61.4 & 57.6 & 42.4 & 10.8 & 3.1 \\
Qwen3.5-9B                             & 55.2 & 49.5 & 34.2 & 11.7 & 1.9 & 58.0 & 52.9 & 35.7 & 12.3 & 3.7 \\
Qwen3.5-4B                             & 48.7 & 40.1 & 29.6 & 1.9  & 0.0 & 47.4 & 40.4 & 27.4 & 1.9  & 0.6 \\
\bottomrule
\end{tabular}
\caption{
  Performance on \bench{} across 1{,}296 turns over 324 clinical sessions
  (4 scenarios $\times$ 3 data richness tiers $\times$ 27 sessions each).
  \textit{mRS}: mean rubric score (\%) over all 1{,}296 turns.
  \textit{Pass@Turn (@T)}: the fraction of turns with rubric score
    $\geq \tau$ (\%).
  \textit{Pass@Session (@S)}: the fraction of complete 4-turn sessions
    where every turn passes threshold $\tau$ (\%). Numbers highlighted in {\color{red} red} and {\color{blue} blue} denote the best and second-best results, respectively.
}
\label{tab:main_results}
\vspace{-10px}
\end{table*}
Table~\ref{tab:main_results} reports performance of 14 LLMs on \bench{}.
Closed-source and open-weight models perform comparably at the top end.
GLM-5 achieves the highest mRS among all models (69.4 EN, 71.5 ZH), narrowly outperforming Claude-Opus-4.7 (68.3 EN, 69.9 ZH) on this metric. However, Pass@Session tells a different story: Claude-Opus-4.7 leads all models on session-level consistency, reaching 23.5\% (EN) and 26.9\% (ZH) at $\tau{=}.60$, and 8.0\% (EN) and 9.0\% (ZH) at $\tau{=}.75$.
This gap between turn-level and session-level rankings highlights that sustaining quality across all four turns within a session is a distinct capability, not captured by mean rubric score alone.
Within the Qwen3.5 family, mRS increases with model size (48.7→55.2→58.3→66.3 from 4B to 35B), with the 4B model recording 0.0\% Pass@Session at $\tau{=}.75$.
Scenario and data-richness-level breakdowns are reported in Appendix~\ref{sec:full-results}

\paragraph{Implicitness Type and Task Type: Heatmap Analysis}

\begin{figure*}
    \centering
    \includegraphics[width=0.95\linewidth]{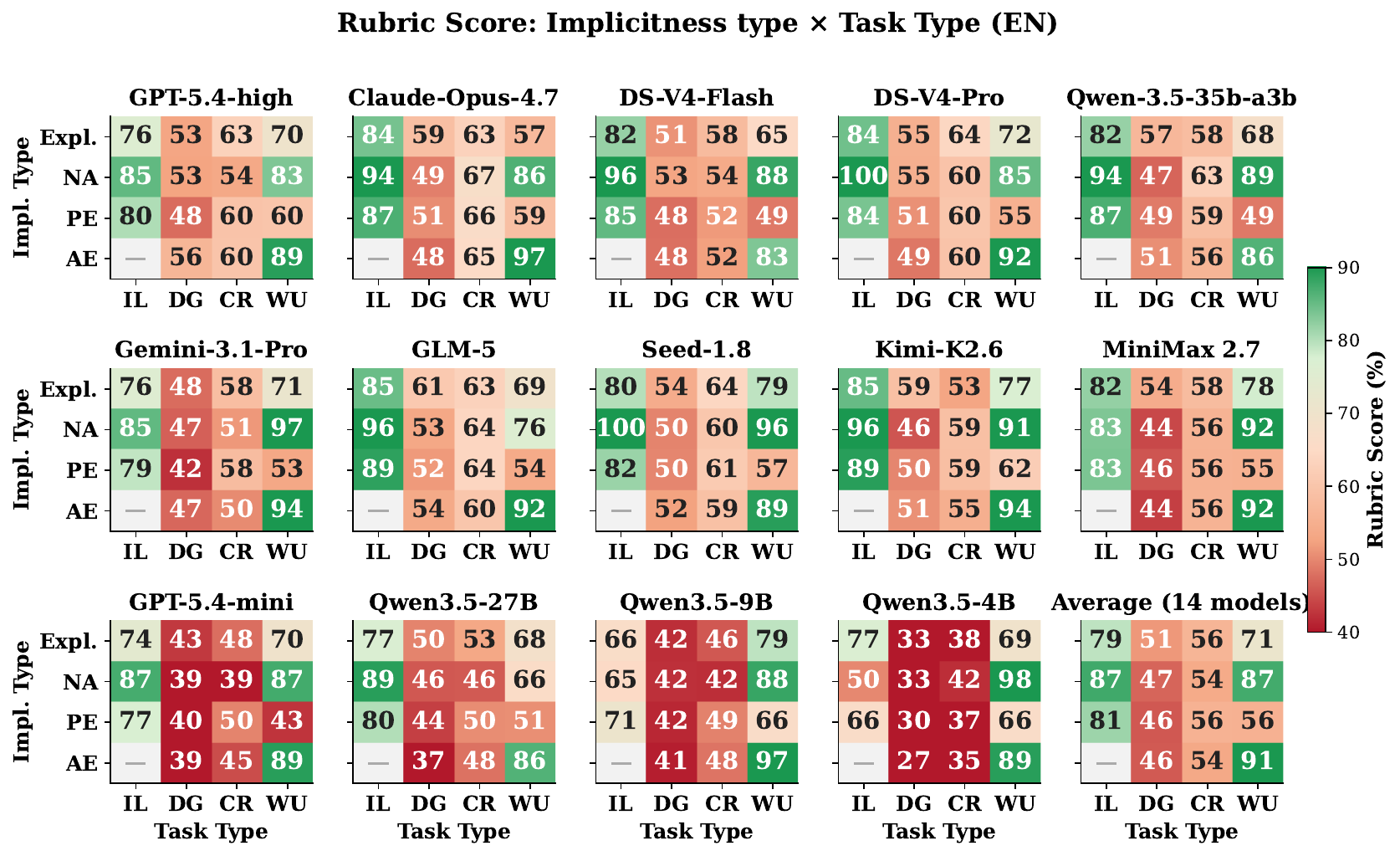}
\caption{Rubric score (\%) by implicitness type $\times$ task type (EN),
for 14 models and their average. Information Lookup stays robust across
all implicitness types, while Data Gathering (multi-tool composition) and
Clinical Reasoning (knowledge-grounded inference) are the consistent weak
points across models. Both demand reasoning beyond single-point retrieval. Results for Chinese (ZH) are presented in Appendix \ref{fig:heatzh}}
\label{fig:heat_grid_en}
\end{figure*}

Figure~\ref{fig:heat_grid_en} decomposes the rubric score by implicitness (rows) and task (columns) across all 14 models. The heatmap 
reveals three patterns. First, task type defines a
stable difficulty hierarchy: IL\,82.6 $>$ WU\,76.1 $>$ CR\,54.9 $>$
DG\,47.5, with DG as the universal bottleneck whose narrow cross-model
standard deviation (5.6--7.6) indicates that multi-tool composition
resists model scale. Second, implicitness interacts non-uniformly with
task type: PE is the weakest row (59.6) but collapses specifically on
WU (55.6), where dropping the verb creates syntactic ambiguity between
a write instruction and a verification query; NA and AE preserve the
write-verb in the antecedent, recovering to 87.2 and 90.7: the two
highest cells on average. Third, cross-model variance is lowest on
AE\,$\times$\,WU ($\sigma{=}4.0$) and highest on NA\,$\times$\,IL
($\sigma{=}13.6$), concentrating the capability gap in implicit DG and
CR cells; Claude-Opus-4.7 and GLM-5 are the most balanced systems, while Qwen3.5-4B shows
the widest within-model spread (98 on NA\,$\times$\,WU vs.\ 27 on
AE\,$\times$\,DG), suggesting surface-pattern acquisition without
underlying composition skill.

\paragraph{Language-Conditioned Tool-Invocation Bias.}
All models trained by non-Chinese organizations achieve higher scores on Chinese than English
sessions (up to $+$11.4\,pp Pass@Session for Gemini~3.1~Pro).
A turn-level breakdown for Gemini~3.1~Pro shows that 23\% of turns where
Chinese passed but English failed involve insufficient tool use (the English
session skips EHR calls correctly invoked in Chinese), and 3\% are outright
refusals absent in Chinese entirely. We attribute this to a
language-conditioned safety prior: English training data reinforces
disclaimers about AI systems not accessing clinical records, suppressing
EHR tool calls in English sessions. This motivates reporting multilingual
results separately.


\paragraph{A Narrow Flagship Band.}
Among the flagship models, performance compresses into a narrow
band: the top contenders (GPT-5.4-high, GLM-5, Claude-Opus-4.7, Kimi-K2.6,
Seed-1.8) fall within \textasciitilde 5 points on EN despite differences in scale and architecture. We attribute this to the information filtering inherent in the data construction pipeline: since EHR evidence is fetched on demand, parametric knowledge differences are removed at the data-construction stage, so models differ in how they \emph{compose} tool calls, not in what they \emph{know}. Lightweight models stay clearly behind, with Qwen3.5-4B trailing GLM-5 (EN 69.4) by \textasciitilde 20 points. As retrieval already removed the knowledge axis, this gap isolates a reasoning deficit: smaller models obtain the same evidence but fail to reliably chain tool calls and integrate the results.

\paragraph{Invariance Across Data Richness Tiers.}
Performance is broadly invariant to the L1--L3 stratification.
Across evaluated models (Appendix \ref{tab:difficulty_breakdown}), mRS varies by less than
\textasciitilde 4 points between L1 and L3, and the ranking is not
monotonic for any model (L3 exceeds L1, e.g.,
Seed-1.8 EN, Qwen3.5-35B EN).
This is a positive finding: under the FHIR tool abstraction,
EHR scale is not the dominant bottleneck.
Models navigate information-dense encounters via targeted tool calls
just as effectively as sparse ones; the dominant axes of variance are
instead task type and implicit-query
type (Figure \ref{fig:heat_grid_en}).

\subsection{Effect of Implicit Queries}
\begin{table}[!htbp]
\centering\footnotesize
\setlength{\tabcolsep}{4pt}
\renewcommand{\arraystretch}{1.15}
\resizebox{\columnwidth}{!}{%
\begin{tabular}{@{}l ccc | ccc | ccc@{}}
\toprule
\multirow{2}{*}{\textbf{Model}}
  & \multicolumn{3}{c|}{\textbf{\NA{}}}
  & \multicolumn{3}{c|}{\textbf{\PE{}}}
  & \multicolumn{3}{c}{\textbf{\AAE{}}} \\
\cmidrule(lr){2-4} \cmidrule(lr){5-7} \cmidrule(l){8-10}
  & Expl. & Impl. & \textbf{$\Delta$}
  & Expl. & Impl. & \textbf{$\Delta$}
  & Expl. & Impl. & \textbf{$\Delta$} \\
\midrule
Claude-Opus-4.7    & 63.2 & 60.8 & $-$2.4          & 67.0 & 65.7 & $-$1.3          & 63.8 & 58.1 & $-$5.7 \\
GPT-5.4-high       & 60.3 & 59.5 & $-$0.8          & 64.7 & 60.7 & $-$4.0          & 61.5 & 58.7 & $-$2.8 \\
Gemini-3.1-Pro     & 56.2 & 54.9 & $-$1.3          & 57.7 & 57.7 & $+$0.0          & 55.2 & 48.8 & $-$6.4 \\
DeepSeek-V4-Pro    & 62.8 & 61.7 & $-$1.1          & 62.8 & 63.0 & $+$0.2          & 58.2 & 55.5 & $-$2.7 \\
DeepSeek-V4-Flash  & 65.8 & 59.9 & $-$5.9          & 64.1 & 59.4 & \textbf{$-$4.7} & 60.1 & 50.2 & \textbf{$-$9.9} \\
Seed-1.8    & 64.0 & 63.2 & $-$0.8          & 67.7 & 65.7 & $-$2.0          & 60.0 & 55.7 & $-$4.3 \\
Qwen3.5-35B-A3B    & 59.1 & 59.2 & $+$0.1          & 64.9 & 62.5 & $-$2.4          & 57.1 & 54.0 & $-$3.1 \\
Qwen3.5-27B        & 57.8 & 48.0 & \textbf{$-$9.8} & 59.2 & 55.9 & $-$3.3          & 51.3 & 44.1 & $-$7.2 \\
\midrule
\textbf{Average}   & 61.2 & 58.4 & \textbf{$-$2.8} & 63.5 & 61.3 & \textbf{$-$2.2} & 58.4 & 53.1 & \textbf{$-$5.3} \\
\bottomrule
\end{tabular}
}
\caption{{Explicit vs.\ implicit ablation (per subtype).} Each model is
  evaluated twice over the entire benchmark: once with every query
  restored to a \emph{fully explicit} paraphrase (\textit{Expl.}), and
  once with the original \emph{implicit} physician queries
  (\textit{Impl.}); all other inputs (patient data, tools, gold
  answers) are held fixed.
  $\Delta = \textit{Impl.} - \textit{Expl.}$ is the performance lost to
  implicit phrasing (negative = implicitness hurts).
  Columns report the comparison restricted to turns of each implicit subtype
  (\NA{}, \PE{}, \AAE{}). The magnitude of $\Delta$ consistently grows
  with subtype difficulty (\PE{} $<$ \NA{} $<$ \AAE{}). All are
  rubric scores (\%).}
\label{tab:results-implicit}

\vspace{-10px}

\end{table}

\bench{} preserves both implicit and explicit versions of each query,
allowing us to isolate the effect of implicit phrasing while keeping
patient data, tools, and gold answers fixed.
As shown in Table~\ref{tab:results-implicit}. The implicit penalty $\Delta$ is negative in 21 of 24
model$\times$subtype cells, showing that implicitness introduces a real
difficulty rather than annotation noise: making queries explicit almost
always improves accuracy.
The penalty increases with anaphoric complexity, averaging $-2.2$ for
Predicate Ellipsis, $-2.8$ for Nominal Anaphora, and $-5.3$ for Abstract
Event Anaphora (\AAE{}), the only subtype where every model degrades.
This ordering is intuitive: predicate ellipsis omits a locally
recoverable verb, nominal anaphora refers to a specific entity, while
\AAE{} requires reconstructing an accumulated clinical state across
multiple turns.
Implicit penalties also separate models more clearly than raw accuracy.
Stronger systems (e.g., DeepSeek-V4-Pro, GPT-5.4-high) remain within
$-4.0$ across all subtypes, whereas smaller open models are much more
fragile (e.g., Qwen3.5-27B: $-9.8$ on \NA{}; DeepSeek-V4-Flash:
$-9.9$ on \AAE{}). Since 
real physicians do not usually restate full context,
this explicit-implicit gap highlights a fundamental
weakness of current EHR assistants in everyday clinical dialogue.

\subsection{Effect of Patient Communication}


\begin{figure}[!htbp]
    \centering
    \includegraphics[width=\linewidth,keepaspectratio]{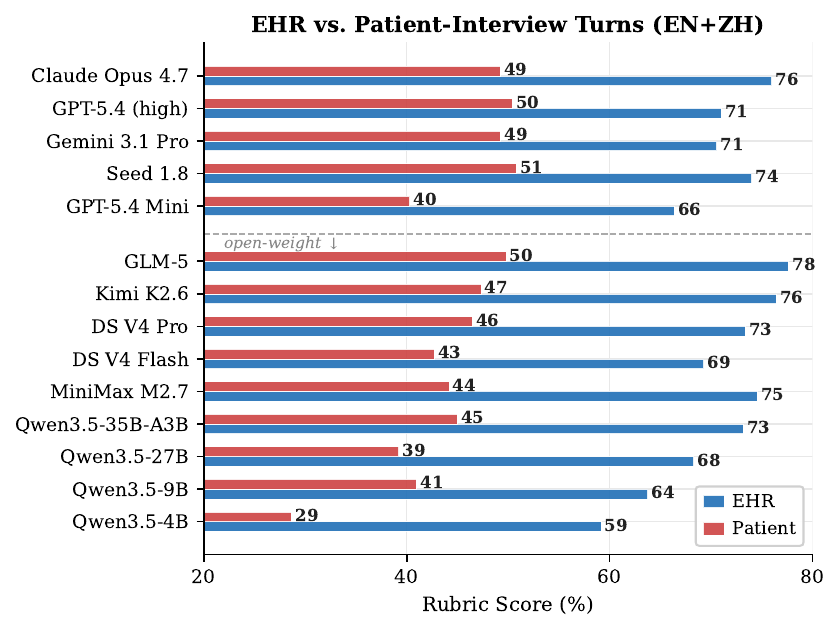}
    \caption{The comparison of rubric scores between EHR-only and Patient-interview turns.}
    \label{fig:ehr-vs-pi}
    \vspace{-10px}
\end{figure}

Figure~\ref{fig:ehr-vs-pi} compares model performance on structured
EHR-access turns versus patient-interview turns.
Across the benchmark (EN+ZH), each model is evaluated on approximately
1{,}944 EHR-only turns and 648 patient-interview turns.
All fourteen models show a substantial drop on patient-interview turns,
with an average decrease of 26.4 points.
EHR-turn scores range from 59\% to 78\%, while patient-interview scores
drop to 29\%--51\%.

The gap reflects a key difference in task structure.
EHR turns mainly require retrieving or filtering well-defined records,
where the correct tool usage is largely determined by the query.
In contrast, patient-interview turns require models to elicit subjective
information through natural dialogue and determine medically relevant
follow-up questions, combining conversational ability with clinical reasoning.
The gap is slightly smaller for stronger proprietary models
(20.6--26.8 points) than for smaller open-weight models
(22.8--30.5 points), but remains substantial across all models. While 
AgentClinic~\cite{schmidgall2026agentclinic} evaluates doctor-patient dialogue 
only, our results show that EHR access does not reduce this difficulty; integrating 
both sources only compounds the challenge. This gap persists uniformly across 
all models, suggesting that patient communication is a capability bottleneck 
independent of general model capacity.


\section{Conclusion}
\label{sec:conclusion}


We introduced \bench{}, the first benchmark that jointly evaluates LLMs on three co-occurring challenges of physician-EHR interaction: implicit physician queries, structured FHIR-based EHR tool use, and ambiguous patient communication.
Built from real MIMIC-IV cases, \bench{} uses a scalable multi-agent synthetic data pipeline to construct \textit{agentic patients} for interactive doctor-patient-EHR scenarios.
Our results show that even the strongest LLMs remain far from reliable as physician assistants, highlighting interaction and coordination as key bottlenecks for clinical LLMs.
We release the benchmark and code to support research on LLM clinical deployments.



\section*{Limitations}

\paragraph{Scenario Coverage.}

\bench{} spans four clinical scenarios chosen to cover a representative range of EHR reasoning and tool-use demands.
Real-world clinical workflows include many additional contexts, such as critical-care management, specialist referral, and post-discharge follow-up, that fall outside the current scope.
We plan to extend the scenario set in future releases, especially to higher-stakes and time-sensitive settings.



\paragraph{Model Coverage and Task Depth.}

\bench{} is designed to stress-test multi-turn reasoning, structured tool use, and implicit query understanding rather than biomedical knowledge recall alone.
We therefore focus on frontier general-purpose LLMs with strong tool-use capabilities.
Domain-specific medical LLMs, such as Med-PaLM~2 and HuatuoGPT, are important baselines, but many are primarily optimized for biomedical QA or clinical knowledge tasks, are not publicly accessible, or are not readily adapted to FHIR-style tool interfaces.
Evaluating medical specialist models with comparable tool-use scaffolds is an important direction for future work.

\paragraph{Agent Design Generality.}
Although \textit{agentic patients} are grounded in real records, they remain tailored to predefined scenarios and task types.
In principle, static patient records could be transformed into richer interactive environments beyond dialogue, including more complex simulations such as in-silico clinical trials.
Our current pipeline, however, uses a manually designed agent workflow for doctor-patient-EHR assistance and does not yet include a meta-pipeline for automatically designing new agent structures for substantially different scenarios.
Developing such adaptive agent-pipeline design is an interesting direction for future work.

\paragraph{Data Source.}


All patients are drawn from MIMIC-IV, a US academic medical dataset with a heavy ICU focus, which may not generalize to documentation styles or disease prevalence in other healthcare systems.

\paragraph{Language.}
\bench{} is currently available in English and Chinese; supporting additional languages would require re-grounding clinical terminology and patient persona behaviour in language-specific medical corpora.


\section*{Acknowledgments}

This study was supported by ELLIS Institute Finland and
School of Electrical Engineering, Aalto University. We acknowledge CSC-IT Center
for Science, Finland, for providing access to the supercomputers Roihu as well as LUMI, owned by the European High Performance Computing
Joint Undertaking (EuroHPC JU) and hosted by CSC Finland in collaboration
with the LUMI consortium. We also acknowledge the computational resources
provided by the Aalto Science-IT project through the Triton cluster.

\bibliography{hintbib/string,hintbib/custom}

\appendix

\renewcommand{\thetable}{\thesection\arabic{table}}
\renewcommand{\thefigure}{\thesection\arabic{figure}}



\section{Data, Licensing, and Ethical Considerations}
\label{sec:ethics-appendix}

\subsection{Data Licence and Distribution}
\label{sec:license}

\bench{} is built on MIMIC-IV~\cite{johnson2023mimic}, released on
PhysioNet under the \textbf{PhysioNet Credentialed Health Data
Licence~1.5.0}.
Access requires completing a recognised human-subjects research
training programme (e.g.\ CITI ``Data or Specimens Only Research''
course) and signing a data-use agreement on PhysioNet.
Because every \bench{} session is a derivative of MIMIC-IV patient
records, the benchmark session files inherit the same licence
conditions and \textbf{may not be redistributed publicly}.

We release all artefacts independently under separate terms:
\begin{itemize}[nosep,leftmargin=1.2em]
  \item \textbf{Code and evaluation harness} (tool API, judge prompts,
        scoring scripts): released under the Apache License 2.0.
  \item \textbf{Benchmark session files} (JSONL containing
        patient-derived data): distributed through a PhysioNet-linked
        repository; requesters must hold an active MIMIC-IV data-use
        agreement before access is granted.
  \item \textbf{Pre-computed evaluation results} (model outputs, rubric
        scores): released openly as they contain no patient-level data.
\end{itemize}

\subsection{Consistency with Intended Use}
\label{sec:intended-use}

MIMIC-IV was created to support clinical and translational research,
quality improvement, and the development of clinical decision-support
tools~\cite{johnson2023mimic}.
Constructing a benchmark to evaluate LLMs on physician--EHR
interaction is fully consistent with that intent.

We impose the following restrictions on downstream use of \bench{}:
\begin{itemize}[nosep,leftmargin=1.2em]
  \item The benchmark is intended for \emph{research and evaluation
        only} and must not be deployed in real clinical workflows.
  \item Benchmark sessions must not be used as training data for
        evaluated models, to prevent leaderboard contamination.
  \item Derivative works must respect the original PhysioNet access
        conditions and must not be used outside research contexts.
\end{itemize}

\subsection{Privacy and De-identification}
\label{sec:privacy}

MIMIC-IV is de-identified by the MIT Laboratory for Computational
Physiology following HIPAA Safe Harbor standards~\cite{johnson2023mimic}:

\begin{itemize}[nosep,leftmargin=1.2em]
  \item \textbf{Direct identifiers removed:} patient names, geographic
        subdivisions finer than state, telephone numbers, and device
        identifiers are absent from the released data.
  \item \textbf{Date shifting:} all dates are offset by a random,
        per-patient amount anchored to the patient's
        \texttt{anchor\_year}, making true calendar dates
        unrecoverable.
  \item \textbf{Age capping:} patients older than 89 are grouped into
        a single 90$+$ anchor-year bin to prevent age-based
        re-identification.
\end{itemize}

\bench{} preserves all of these protections and introduces no
additional identifiers.
Session files contain MIMIC-IV \texttt{subject\_id} values, which are
surrogate keys with no mapping to real-world identities outside the
MIMIC-IV access-controlled environment.
Clinical note content reproduced in sessions (discharge summaries,
radiology reports) retains no names or explicit dates.

Despite these protections, re-identification risk is non-zero:
discharge summaries may describe unusual disease combinations that,
combined with external auxiliary information, could narrow a patient's
identity.
We therefore require credentialed access for all session files,
consistent with PhysioNet terms.

We additionally note that MIMIC-IV clinical text reflects real
physician language, which may contain implicit demographic biases or
culturally specific clinical framing.
Systematic content auditing beyond the PhysioNet de-identification
pipeline was not performed; this is acknowledged as a limitation
in \S\ref{sec:conclusion}.

\subsection{Artefact Documentation and Coverage}
\label{sec:artifact-doc}

\paragraph{Domain and source.}
All patient data originate from Beth Israel Deaconess Medical Center
(BIDMC), Boston, MA, USA, collected approximately 2008--2019.
BIDMC is a large academic tertiary-care centre with a heavy ICU and
internal-medicine case mix.
Benchmark performance may not generalise to community hospitals,
non-US healthcare systems, or populations with substantially
different disease prevalence.

\paragraph{Languages.}
All sessions are provided in English and Mandarin Chinese.
English queries are generated directly from MIMIC-IV data; Chinese
queries are produced via a round-trip translation pipeline with
semantic verification.
Underlying clinical notes (discharge summaries, radiology reports)
remain in English throughout, as MIMIC-IV contains no Chinese-language
source records.
Patient-simulation responses are available in both languages.

\subsection{Dataset Statistics}
\label{sec:checklist-stats}

\paragraph{Dataset statistics.}
The complete benchmark composition, including scenario, data-richness,
task-type, and implicitness distributions, is reported in
Appendix~\ref{sec:full-stats}.

\subsection{Use of AI Assistance}
\label{sec:ai-assistance}

During manuscript preparation, we used LLMs only for grammar correction,
language refinement, and limited assistance in literature search. All cited
references were manually checked and verified by the authors. The authors
reviewed and edited all AI-assisted text and take full responsibility for
the content of the manuscript.

Other uses of LLMs are part of the scientific design of this work and are
explicitly described in the relevant sections. First, LLMs are the evaluated
models in all experiments (Section~\ref{sec:experiments}). Second, LLMs are
used as components of the multi-agent benchmark construction pipeline,
including the Session Planner, Doctor Agent, Patient Agent, and quality
checker (Section~\ref{sec:benchmark}). Third, GPT-5.4-mini is used as the
automated rubric judge for turn-level scoring (Section~\ref{sec:eval}), with
human agreement validation reported in
Appendix~\ref{sec:judge-reliability}.

\subsection{Ethical Considerations}
\bench{} is built from MIMIC-IV, a publicly available de-identified clinical dataset, and we followed its standard data-use and access requirements.
The benchmark is intended only for research evaluation, not for clinical deployment or autonomous decision-making.
As a secondary-use benchmark with synthetic interactions, \bench{} may inherit biases from the source records and introduce additional biases through scenario planning, patient personas, LLM-generated dialogue, and record filtering.
In particular, because the pipeline selects records that can support predefined scenarios, it may introduce implicit cohort-selection effects, which should be considered for clinically sensitive evaluations.
We mitigate these risks through record grounding, filtering, and manual review, but the benchmark should still be interpreted as an evaluation resource rather than a representation of real patient experience.
\section{Data and Code Availability}
\label{sec:availability}

\setcounter{table}{0}
\setcounter{figure}{0}

Our dataset is derived from MIMIC-IV~\cite{johnson2023mimic} under the
PhysioNet Credentialed Health Data License. Following the
re-distribution requirements of the original license, the dataset will be released formally via PhysioNet. Access requires a valid PhysioNet credentialed account and acceptance of the corresponding data use agreement. The released dataset includes: (1) de-identified
EHR snapshots extracted from MIMIC-IV admissions used as session
inputs; (2) per-subject patient records partitioned by \texttt{subject\_id},
containing structured clinical data (diagnoses, medications,
observations, procedures); (3) 324 annotated clinical sessions 1296 turns in both
English and Chinese; (4) gold-standard tool-call trajectories and reference
answers for each turn; and (5) evaluation rubrics per turn
with per-item pass/fail annotations.
All protected health information (PHI) has been removed.

The codebase is publicly available on GitHub under the Apache License 2.0. 
It includes scripts to extract and partition raw MIMIC-IV records by \texttt{subject\_id}, converting them into EHR snapshots, running the scalable multi-agent synthetic data pipeline for constructing \textit{agentic patients} (Section~\ref{sec:agentic_pipeline}) and reproducing reported evaluations. It also provides sample data for setup without requiring full MIMIC-IV access.

\section{Human Evaluation Details}
\label{sec:human-eval}
\setcounter{table}{0}
\setcounter{figure}{0}

\subsection{Expert Reviewer Backgrounds}
\label{sec:reviewer-backgrounds}
\paragraph{Clinical Expert.}
We recruited one volunteer senior physician (female), board-certified
in her country of practice and with 15 years of clinical experience.
Participation was voluntary and uncompensated (no monetary reward).
The participant used English and Chinese as the language of instruction.

\paragraph{Trained Annotators.}
We recruited 8 volunteer trained annotators (two female,
six male), each with several years of clinical AI and NLP research experience. Those with more than 8 years of NLP research experience are regarded as NLP experts.
Participation was voluntary and uncompensated (no monetary reward).
The participants used English and Chinese as the languages of
instruction.

\subsection{Human Review and Field Verification}
\label{sec:review-process}
The physician review and trained-annotator review served distinct purposes.
The former assessed session-level clinical plausibility, whereas the latter
performed field-level verification and correction across the released data.

\paragraph{Stage 1: Clinical Plausibility Review.}
After the pipeline produced the initial dataset, we sampled 25\% of
the sessions for clinical review.
The clinical expert assessed each sampled dialogue for clinical
plausibility, i.e., whether the interaction could realistically occur
in practice and whether the content contained any internal medical
contradictions.
On the sampled subset, 95\% of sessions were judged clinically
plausible.
Two NLP experts then collected the sessions flagged as implausible,
distilled the clinical expert's comments into additional pipeline
prompt constraints, and regenerated those sessions.

\paragraph{Stage 2: Field-Level Verification.}
In the second stage, all 8 trained annotators verified the correctness
of each annotated field of every session, covering both the
previously approved sessions and the regenerated ones.
Three fields were checked: (i) the implicitness type, i.e., whether
the assigned type matches the actual implicit query; (ii) the rubric,
i.e., whether each item corresponds one-to-one with its gold answer;
and (iii) the gold tool calls, i.e., whether their parameters are
correct.
The implicitness-type assignment matched the query in 87\% of cases,
and only 8\% of rubric items required revision.
The trained annotators reviewed and corrected all flagged fields, yielding 100\% turn-level coverage. 

\subsection{Full-Benchmark Realism Assessment}
\label{sec:realism-assessment}

The board-certified physician and the trained annotators rated each
session along two dimensions: \textit{query realism}, whether the physician's
query resembles a request that could occur in practice, and \textit{trajectory
coherence}, whether the four turns form a clear clinical workflow. Each
dimension was scored from 1 (lowest) to 5 (highest); scores below 3 were
designated as low quality. The session-level realism score is the mean of the
two dimension scores.

\begin{table}[t]
\centering
\small
\setlength{\tabcolsep}{3.5pt}
\begin{tabular}{lrrrr}
\toprule
\textbf{Scenario} & \textbf{$n$} & \textbf{Mean} & \textbf{Min.} & \textbf{$\geq4.5$} \\
\midrule
Diagnostic Workup  & 81 & 4.42 & 4.00 & 73\% \\
Medication Safety  & 81 & 4.47 & 4.00 & 79\% \\
Treatment Response & 81 & 4.39 & 4.00 & 78\% \\
Discharge Planning & 81 & 4.20 & 3.50 & 47\% \\
\midrule
\textbf{Overall}   & \textbf{324} & \textbf{4.37} & \textbf{3.50} & \textbf{69\%} \\
\bottomrule
\end{tabular}
\caption{Session-level realism ratings by clinical scenario. The final column
reports the percentage of sessions with a mean score of at least 4.5.}
\label{tab:realism-by-scenario}
\end{table}

\begin{table}[t]
\centering
\footnotesize
\setlength{\tabcolsep}{4pt}
\begin{tabular}{lrr}
\toprule
\textbf{Scenario} & \textbf{Query realism} & \textbf{Trajectory}\\
\midrule
Diagnostic Workup  & 4.33 & 4.51 \\
Medication Safety  & 4.28 & 4.65 \\
Treatment Response & 3.80 & 4.98 \\
Discharge Planning & 3.70 & 4.69 \\
\midrule
\textbf{Overall}   & \textbf{4.03} & \textbf{4.71} \\
\bottomrule
\end{tabular}
\caption{Mean scores for the two realism dimensions.}
\label{tab:realism-by-dimension}
\end{table}

The overall mean is 4.37/5, and no session falls below the quality
threshold. Table~\ref{tab:lowest-realism-session} shows the lowest-scoring
session. Its mean score of 3.5 reflects the textbook-like wording of Turn~1,
rather than a clinical contradiction; the four turns still form a coherent
workflow.

\newcolumntype{L}[1]{>{\raggedright\arraybackslash}p{#1}}
\newcolumntype{Y}{>{\raggedright\arraybackslash}X}

\begin{table}[ht]
\centering
\footnotesize
\setlength{\tabcolsep}{3pt}
\renewcommand{\arraystretch}{1.08}

\begin{tabularx}{\columnwidth}
  {@{}l >{\raggedright\arraybackslash}p{1.65cm} c Y@{}}
\toprule
\textbf{Turn}
& \textbf{Task type}
& \textbf{Subtype}
& \textbf{Physician query} \\
\midrule

T0
& Information Lookup
& Explicit
& What was the reason for this admission? \\
\addlinespace[2pt]

T1
& Data Gathering
& AE
& In summary, what is the management plan for catheter-related infection in patients receiving immunosuppressive drugs? \\
\addlinespace[2pt]

T2
& Data Gathering
& PE
& Is the patient taking Tacrolimus as prescribed, and how are they managing daily activities at home? \\
\addlinespace[2pt]

T3
& Write/\newline Update
& NA
& Add the medication I just mentioned, 2 mg orally twice daily, to the medication list, as the patient reported not taking it and is currently using other immunosuppressants. \\

\bottomrule
\end{tabularx}

\caption{The lowest-scoring session
(\texttt{discharge\_planning\_15}): query realism 3, trajectory
coherence 4, and mean realism 3.5.}
\label{tab:lowest-realism-session}
\end{table}

\subsection{Judge Model Reliability}
\label{sec:judge-reliability}

We evaluated the judge-agreement study on 108 sessions per language, one third of the 324-session benchmark in each language. Using GLM-5 responses as the
scoring target, the eight trained annotators scored the same rubric items as
the fixed GPT-5.4-mini judge. The resulting comparison contains 3,864 rubric
items. The judge and trained-annotator labels agree on 3,685 items, yielding
95.37\% agreement. A two-sided normal-approximation binomial confidence
interval, $\hat p \pm 1.96\sqrt{\hat p(1-\hat p)/n}$, has a half-width of
0.66 percentage points.

\begin{table}[ht]
\centering
\footnotesize
\setlength{\tabcolsep}{2.5pt}
\begin{tabular}{lrrrr}
\toprule
 & \textbf{\IL{}} & \textbf{\DG{}} & \textbf{\CR{}} & \textbf{\WU{}} \\
\midrule
EN agreement       & 0.9621 & 0.9518 & 0.9605 & 0.9782 \\
ZH agreement       & 0.9431 & 0.9419 & 0.9542 & 0.9455 \\
EN disagreements   & 16 & 29 & 25 & 6 \\
ZH disagreements   & 24 & 35 & 29 & 15 \\
Total items        & 844 & 1,204 & 1,266 & 550 \\
\bottomrule
\end{tabular}
\caption{Judge agreement by task type and language. Total-item counts combine English and Chinese; each language contributes half of the items in every task column.}
\label{tab:judge-agreement-task}
\end{table}

\begin{table}[ht]
\centering
\footnotesize
\setlength{\tabcolsep}{4pt}
\begin{tabular}{lrrrr}
\toprule
 & \textbf{Factual} & \textbf{Clinical} & \textbf{Context} & \textbf{Safety} \\
\midrule
EN agreement       & 0.9654 & 0.9596 & 0.9355 & 0.9641 \\
ZH agreement       & 0.9578 & 0.9358 & 0.9113 & 0.9521 \\
EN disagreements   & 32 & 29 & 8 & 6 \\
ZH disagreements   & 39 & 46 & 11 & 8 \\
Total items        & 1,848 & 1,434 & 248 & 334 \\
\bottomrule
\end{tabular}
\captionof{table}{Judge agreement by rubric category and language: factual accuracy,
clinical interpretation, context integration, and hallucination/safety.
Total-item counts combine English and Chinese.}
\label{tab:judge-agreement-category}
\end{table}

Agreement exceeds 91\% in every reported task-, language-, and
rubric-category subset. This study evaluates agreement on one target model
(GLM-5) and should therefore be interpreted as evidence for judge reliability
on the sampled outputs, rather than exhaustive validation across all model
output distributions.
\setcounter{table}{0}
\setcounter{figure}{0}

\section{EHR Tool Inventory}
\addcontentsline{toc}{section}{Appendix A: EHR Tool Inventory}
\label{app:tools}

\setcounter{table}{0}
\setcounter{figure}{0}

\bench{} exposes \textbf{18 tools} to the evaluated agent, organized into four groups: EHR read tools, EHR write tools, patient-interview tools,
and one control tool.
All EHR tools follow the FHIR R4 naming convention
(\texttt{ResourceType.operation}); patient-interview tools use the
\texttt{patient.*} namespace. Each tool call requires a \texttt{subject\_id} (MIMIC-IV patient identifier); most read tools also accept an optional \texttt{hadm\_id} to scope results to
a single admission.

\newcommand{\tool}[1]{\texttt{\small #1}}

\setlength{\tabcolsep}{4pt}
\renewcommand{\arraystretch}{1.12}

\subsection{EHR Read Tools (9)}
\label{app:tools:read}

\begin{table*}[!htbp]
\centering
\footnotesize
\begin{tabular}{@{}p{4.9cm} p{7.8cm} p{2.6cm}@{}}
\toprule
\textbf{Tool name} & \textbf{Description} & \textbf{Key opt.\ params} \\
\midrule

\tool{Patient.read}
  & Returns demographic information (gender, anchor age, deceased date).
  & — \\[2pt]

\tool{Encounter.search}
  & Retrieves all hospital admissions with admission/discharge times,
    admission type, ward, and discharge disposition.
  & \texttt{hadm\_id} \\[2pt]

\tool{Condition.search}
  & Returns ICD diagnosis conditions ordered by \texttt{seq\_num}
    (principal diagnosis first).
    Supports filtering by ICD code prefix (e.g.\ \texttt{K74}) or
    clinical status (\textit{active / resolved / inactive}).
  & \texttt{hadm\_id}, \texttt{code}, \texttt{clinical\_status} \\[2pt]

\tool{Observation.search}
  & The primary data-retrieval tool.
    Covers laboratory results (\texttt{category=laboratory}),
    ICU vital signs (\texttt{category=vital-signs}), and
    microbiology culture results (\texttt{category=microbiology}).
    The \texttt{code} parameter accepts a free-text lab/vital name
    (e.g.\ \textit{Creatinine}, \textit{Heart Rate}); results are
    sorted newest-first.
  & \texttt{hadm\_id}, \texttt{category}, \texttt{code},
    \texttt{date\_from}, \texttt{date\_to}, \texttt{\_count} \\[2pt]

\tool{MedicationRequest.search}
  & Returns prescription (medication order) records sorted by start
    time newest-first.
    Accepts drug-name substring filter and order-status filter
    (\textit{active / completed / stopped}).
  & \texttt{hadm\_id}, \texttt{medication}, \texttt{status} \\[2pt]

\tool{MedicationAdministration.search}
  & Returns eMAR records showing which medications were
    \emph{actually administered}, when, and at what dose---distinct
    from what was prescribed.
  & \texttt{hadm\_id}, \texttt{medication} \\[2pt]

\tool{DiagnosticReport.search}
  & Returns radiology and other diagnostic reports with full report text.
    Accepts modality filter (\textit{CT, MRI, X-ray, Echo}).
  & \texttt{hadm\_id}, \texttt{report\_type} \\[2pt]

\tool{DocumentReference.search}
  & Searches clinical notes (discharge summaries and radiology reports).
    Supports document-type selection (\textit{discharge-summary /
    radiology}) and full-text keyword search.
  & \texttt{hadm\_id}, \texttt{type\_code}, \texttt{keyword} \\[2pt]

\tool{CarePlan.search}
  & Returns post-discharge care plan sections extracted from MIMIC-IV
    discharge notes: discharge instructions, follow-up schedules, and
    inpatient treatment plans.
    Filter by category (\textit{discharge-planning / followup /
    treatment}).
  & \texttt{hadm\_id}, \texttt{category} \\

\bottomrule
\end{tabular}
\caption{EHR read tools (FHIR R4 naming). All tools require
\texttt{subject\_id}; most accept an optional \texttt{hadm\_id} to scope
results to a single hospital admission.}
\label{tab:tools_read}
\end{table*}
As shown in Table~\ref{tab:tools_read}, these 9 read-only tools expose
MIMIC-IV structured records via a FHIR R4--compatible interface, covering
patient demographics, encounters, diagnoses, observations (lab, vitals,
microbiology), medications, diagnostic reports, clinical notes, and care plans.
\texttt{Observation.search} is the most heavily used tool, unifying
laboratory, vital-sign, and microbiology data under one parameterised call.

\subsection{EHR Write Tools (3)}
\label{app:tools:write}

\begin{table*}[!htbp]
\centering
\footnotesize
\begin{tabular}{@{}p{4.6cm} p{7.0cm} p{3.4cm}@{}}
\toprule
\textbf{Tool name} & \textbf{Description} & \textbf{Required params} \\
\midrule

\tool{MedicationRequest.create}
  & Creates a new medication order (prescription).
    Used for discharge medication additions or inpatient dose changes.
  & \texttt{medication}, \texttt{dose},
    \texttt{route}, \texttt{frequency} \\[2pt]

\tool{ServiceRequest.create}
  & Creates a referral or service request for discharge planning
    (e.g.\ home health, physical therapy, social work,
    specialist follow-up).
  & \texttt{service\_type}, \texttt{priority} \\[2pt]

\tool{Flag.create}
  & Creates a safety-critical alert flag on the patient record
    (e.g.\ high fall risk, allergy update, mandatory follow-up).
  & \texttt{category}, \texttt{code}, \texttt{detail} \\

\bottomrule
\end{tabular}
\caption{EHR write tools. All tools additionally require \texttt{subject\_id}.}
\label{tab:tools_write}
\end{table*}

As shown in Table~\ref{tab:tools_write}, these 3 tools simulate EHR write operations and are used primarily in
\textit{Write/Update} turns (T3) of the Discharge Planning scenario,
and also appear in Diagnostic Workup, Medication Safety, and Treatment
Response turns that require ordering or flagging actions.

\subsection{Patient Interview Tools (5)}
\label{app:tools:patient}

\begin{table*}[!htbp]
\centering
\footnotesize
\begin{tabular}{@{}p{4.9cm} p{9.8cm}@{}}
\toprule
\textbf{Tool name} & \textbf{Description} \\
\midrule

\tool{patient.get\_symptom\_history}
  & Asks the patient to describe their symptom history in OPQRST style
    (onset, provocation, quality, radiation, severity, timing).
    An optional \texttt{query} parameter focuses the response on a
    specific symptom (e.g.\ \textit{abdominal pain}). \\[2pt]

\tool{patient.get\_medication\_adherence}
  & Asks whether the patient is taking a specific medication as prescribed.
    The required \texttt{drug} parameter names the medication of interest.
    For patients with \textit{critical\_withheld} profiles, naming the
    drug explicitly may surface intentionally withheld non-adherence. \\[2pt]

\tool{patient.get\_social\_history}
  & Asks about the patient's social situation: living arrangements,
    occupation, smoking, alcohol use, and support system. \\[2pt]

\tool{patient.get\_functional\_status}
  & Asks about functional capacity and ability to perform activities of
    daily living (ADLs); returns any reported limitations. \\[2pt]

\tool{patient.get\_pain\_assessment}
  & Asks the patient to characterise any pain: location,
    severity (0--10 NRS), character, onset, and radiation. \\

\bottomrule
\end{tabular}
\caption{Patient-interview tools (\texttt{patient.*} namespace). All tools
require \texttt{subject\_id} and \texttt{session\_id}.}
\label{tab:tools_patient}
\end{table*}
As shown in Table~\ref{tab:tools_patient}, patient-interview tools expose a simulated patient agent to the evaluated model.
The patient responses are pre-generated during benchmark construction,
grounded in the same admission records as the corresponding EHR snapshots, and stored with each session.
During evaluation, each tool call returns the corresponding fixed response through deterministic replay; the patient LLM is not invoked online.
All patient tools require \texttt{subject\_id} and \texttt{session\_id}.
\subsection{Control Tool (1)}
\label{app:tools:ctrl}

\begin{table*}[!htbp]
\centering
\footnotesize
\begin{tabular}{@{}p{4.6cm} p{10.2cm}@{}}
\toprule
\textbf{Tool name} & \textbf{Description} \\
\midrule

\tool{prepare\_to\_answer}
  & Signals that sufficient information has been gathered and the agent
    is ready to produce the final answer to the user.
    The optional \texttt{answer\_type} parameter distinguishes
    tool-grounded answers (\texttt{tool}) from knowledge-only responses
    (\texttt{chat}).
    This tool must appear as the final action of every turn;
    omitting it is counted as an incomplete plan. \\

\bottomrule
\end{tabular}
\caption{Control tool. No mandatory parameters.}
\label{tab:tools_ctrl}
\end{table*}
\texttt{prepare\_to\_answer} (Table~\ref{tab:tools_ctrl}) is a mandatory
bookkeeping call that every turn must end with. It separates the \emph{information-gathering} phase from the \emph{answer-generation} phase: the benchmark records whether the agent issues this call, and omitting it is penalised as an incomplete plan regardless of the quality of the final answer.
\subsection{Tool Usage by Scenario}
\label{app:tools:avail}

\begin{table*}[!htbp]
\centering
\footnotesize
\setlength{\tabcolsep}{6pt}
\renewcommand{\arraystretch}{1.08}
\begin{tabular}{lcccc}
\toprule
\textbf{Tool} &
\textbf{Diag.\ Workup} &
\textbf{Discharge Plan} &
\textbf{Med.\ Safety} &
\textbf{Treat.\ Response} \\
\midrule
\multicolumn{5}{l}{\textit{EHR Read}} \\
\midrule
\tool{Patient.read}                      & \cmark & \xmark & \cmark & \xmark \\
\tool{Encounter.search}                  & \xmark & \cmark & \xmark & \xmark \\
\tool{Condition.search}                  & \cmark & \cmark & \cmark & \cmark \\
\tool{Observation.search}                & \cmark & \cmark & \cmark & \cmark \\
\tool{MedicationRequest.search}          & \cmark & \cmark & \cmark & \cmark \\
\tool{MedicationAdministration.search}   & \cmark & \cmark & \cmark & \cmark \\
\tool{DiagnosticReport.search}           & \cmark & \cmark & \xmark & \xmark \\
\tool{DocumentReference.search}          & \xmark & \cmark & \xmark & \xmark \\
\tool{CarePlan.search}                   & \xmark & \cmark & \xmark & \xmark \\
\midrule
\multicolumn{5}{l}{\textit{EHR Write}} \\
\midrule
\tool{MedicationRequest.create}          & \cmark & \cmark & \cmark & \cmark \\
\tool{ServiceRequest.create}             & \cmark & \cmark & \xmark & \cmark \\
\tool{Flag.create}                       & \cmark & \cmark & \cmark & \cmark \\
\midrule
\multicolumn{5}{l}{\textit{Patient Interview}} \\
\midrule
\tool{patient.get\_symptom\_history}     & \cmark & \cmark & \cmark & \cmark \\
\tool{patient.get\_medication\_adherence}& \cmark & \cmark & \cmark & \cmark \\
\tool{patient.get\_social\_history}      & \cmark & \cmark & \cmark & \xmark \\
\tool{patient.get\_functional\_status}   & \xmark & \cmark & \xmark & \xmark \\
\tool{patient.get\_pain\_assessment}     & \xmark & \xmark & \xmark & \cmark \\
\midrule
\multicolumn{5}{l}{\textit{Control}} \\
\midrule
\tool{prepare\_to\_answer}               & \cmark & \cmark & \cmark & \cmark \\
\bottomrule
\end{tabular}
\caption{Tool usage across the four benchmark scenarios.
\cmark~= appears in the gold-standard trajectory of at least one session;
\xmark~= not required by any gold trajectory in that scenario.}
\label{tab:tool_availability}
\end{table*}

Table~\ref{tab:tool_availability} shows which tools appear as gold-standard
actions in at least one session of each benchmark scenario. During \emph{data generation}, the reference trajectories were constructed under scenario-specific constraints: only clinically relevant tools were included in the gold plans to ensure grounded and parsimonious annotations. During \emph{evaluation}, however, all 18 tools are exposed to the model without restriction, so that the benchmark can assess whether the model selects the appropriate tools, avoids unnecessary calls, and handles the full tool vocabulary rather than a curated subset.

\section{Physician Query Implicitness Example}
Table \ref{tab:subtypes} shows representative examples about different physician query implicitness. 
\setcounter{table}{0}
\setcounter{figure}{0}
\begin{table*}[!htbp]
\centering
\footnotesize
\renewcommand{\arraystretch}{1.35}
\setlength{\tabcolsep}{4pt}

\begin{tabular*}{\textwidth}{@{\extracolsep{\fill}} p{2.5cm} >{\raggedright\arraybackslash}p{3.0cm} >{\raggedright\arraybackslash}p{3.5cm} >{\raggedright\arraybackslash}p{3.0cm} >{\raggedright\arraybackslash}p{3.0cm} @{}}
\toprule

\textbf{Implicitness Type }
& \textbf{Linguistic Mechanism}
& \textbf{Prior Context}
& \textbf{Explicit Query}
& \textbf{Implicit Query} \\

\midrule

\textbf{\NA{}} \newline \textit{Nominal Anaphora}
& A pronoun or noun phrase refers back to a specific entity or value introduced earlier.
& Turn~0: ``What is this patient's most recent creatinine?''
  \newline $\rightarrow$ Cr\,=\,1.8\,mg/dL
& ``Is that creatinine value consistent with Stage\,3 CKD?''
& ``Is \textit{it} consistent with Stage\,3 CKD?'' \\

\midrule

\textbf{\PE{}} \newline \textit{Predicate \newline Ellipsis}
& The main verb and its arguments are omitted; only the new focus element is stated.
& Turn~0: ``What is this patient's most recent creatinine?''
  \newline $\rightarrow$ Cr\,=\,1.8\,mg/dL
& ``What is this patient's most recent eGFR?''
& ``\textit{eGFR?}'' \\

\midrule

\textbf{\AAE{}} \newline \textit{Abstract \newline Event Anaphora}
& A phrase such as ``given all this'' refers to the accumulated clinical context across prior turns.
& Turn~0: Cr\,=\,1.8\,mg/dL. \newline
  Turn~1: eGFR\,=\,35\,mL/min. \newline
  Turn~2: metformin 500\,mg BID.
& ``Given the patient's creatinine and eGFR, is it safe to continue the metformin?''
& ``\textit{Given all of this,} is it safe to continue the metformin?'' \\

\bottomrule
\end{tabular*}

\caption{The three physician-query implicitness types in \bench{},
illustrated with a CKD--metformin scenario. In the Prior Context column, ``$\rightarrow$'' denotes the tool output returned by the agent after executing the query. For \AAE{} row, prior turns are summarized directly and the earlier queries are omitted for brevity. Explicit queries are
semantically complete physician requests, whereas implicit queries omit
recoverable information and more closely resemble real clinical
conversation.  \AAE{} is generally more challenging than \NA{} and \PE{}.}

\label{tab:subtypes}
\end{table*}
\label{sec:implicttypes}

\section{Full Session Example}
\label{sec:full-example}
\begin{CJK}{UTF8}{gbsn}
\setcounter{table}{0}
\setcounter{figure}{0}
Figure~\ref{fig:session_a} and \ref{fig:session_b} present two representative full-session examples illustrating the tool calls, explicit and implicit query reformulations, and the corresponding evaluation rubrics.

\definecolor{bgIL}{RGB}{204,222,247}
\definecolor{bgDG}{RGB}{198,237,207}
\definecolor{bgCR}{RGB}{232,210,248}
\definecolor{bgWU}{RGB}{251,225,180}
\definecolor{bgHdrA}{RGB}{174,209,240}
\definecolor{bgHdrB}{RGB}{215,185,240}
\definecolor{bgSubhdr}{RGB}{220,220,220}

\newcommand{\tn}[1]{\texttt{\tiny #1}}
\newcommand{\pat}[1]{{\tiny\textit{``#1''}}}

\newlength{\sCA}\setlength{\sCA}{1.10cm}   
\newlength{\sCB}\setlength{\sCB}{2.05cm}   
\newlength{\sCF}\setlength{\sCF}{2.35cm}   
\newlength{\sCC}\setlength{\sCC}{3.45cm}   
\newlength{\sCD}\setlength{\sCD}{3.35cm}   
\newlength{\sCE}\setlength{\sCE}{2.60cm}   

\newcommand{\sessionlegend}{%
{\tiny
\colorbox{bgIL}{\phantom{x}}\,\IL\,Information Lookup\enspace
\colorbox{bgDG}{\phantom{x}}\,\DG\,Data Gathering\enspace
\colorbox{bgCR}{\phantom{x}}\,\CR\,Clinical Reasoning\enspace
\colorbox{bgWU}{\phantom{x}}\,\WU\,Write/Update\enspace$|$\enspace
\textbf{Impl.:}\enspace\PE\,Predicate Ellipsis\enspace
\NA\,Nominal Anaphora\enspace\AAE\,Abstract Event Anaphora\enspace—\,Explicit}}

\begin{figure*}[!htbp]
\centering
\scriptsize
\setlength{\tabcolsep}{3pt}
\renewcommand{\arraystretch}{1.10}

\begin{tabular}{@{}p{\sCA}p{\sCB}p{\sCF}p{\sCC}p{\sCD}p{\sCE}@{}}
\hline
\multicolumn{6}{@{}l@{}}{\cellcolor{bgHdrA}\strut
  \normalsize\textbf{Session A:} Discharge Planning\enspace|\enspace
  HIV$^{+}$, type-2 diabetes, CKD (eGFR\,$\approx$\,35)}\\
\hline
\rowcolor{bgSubhdr}
\textbf{Type / Impl.} & \textbf{Implicit query} & \textbf{Explicit query} &
\textbf{Tool calls \& patient response} &
\textbf{Model answer} & \textbf{Rubric criteria} \\
\hline

\rowcolor{bgIL}
\textbf{T0} \IL\newline{\tiny ehr\,/\,—} &
What's the latest HIV viral load? &
What's the latest HIV viral load? &
\tn{Observation.search(code=HIV 1 Viral Load)}\newline
\textit{\tiny$\to$ 0.0 log10 copies/mL} &
HIV viral load: 0.0 log10 copies/mL (undetectable). &
{\tiny $\bullet$ States VL = 0.0 log10 copies/mL}\newline
{\tiny $\bullet$ Interprets as undetectable}\newline
{\tiny $\bullet$ No fabricated data} \\
\hline

\rowcolor{bgIL}
\textbf{T1} \IL\newline{\tiny ehr\,/\,\PE} &
Creatinine? &
What's the latest creatinine? &
\tn{Observation.search(code=Creatinine)}\newline
\textit{\tiny$\to$ Cr = 1.2 mg/dL} &
Creatinine: 1.2\,mg/dL (within normal range). &
{\tiny $\bullet$ Cites Cr = 1.2\,mg/dL}\newline
{\tiny $\bullet$ Interprets as within normal range}\newline
{\tiny $\bullet$ Gives a clear conclusion} \\
\hline

\rowcolor{bgDG}
\textbf{T2} \DG\newline{\tiny mixed\,/\,\AAE} &
Given all this, is metformin still appropriate? &
Based on the creatinine value, is metformin still appropriate? &
\tn{Observation.search(Creatinine)}\newline
\textit{\tiny$\to$ Cr = 1.2 mg/dL}\newline
\tn{MedReq.search(Metformin)}\newline
\textit{\tiny$\to$ Metformin 500\,mg order (active)}\newline
\tn{patient.get\_med\_adherence(Metformin)}\newline
\pat{No, I'm actually not taking the Metformin. It makes me so nauseous I stopped a few weeks ago. I never brought it up.} &
{\tiny $\bullet$ Cr 1.2 does not contraindicate metformin on renal grounds\newline
$\bullet$ Patient stopped metformin due to GI nausea, not renal safety\newline
$\bullet$ Barrier is tolerability; suggest dose reduction or extended-release\newline
$\bullet$ Consider SGLT2/GLP-1 if GI effects persist} &
{\tiny $\bullet$ Cites Cr 1.2; renal function does not contraindicate}\newline
{\tiny $\bullet$ Quotes patient stopping metformin for nausea}\newline
{\tiny $\bullet$ Identifies tolerability (not renal) as the barrier}\newline
{\tiny $\bullet$ Recommends dose reduction / XR; alt.\ agent if persists} \\
\hline

\rowcolor{bgWU}
\textbf{T3} \WU\newline{\tiny write\,/\,\NA} &
OK, please create a medication order to reduce it to 250\,mg orally once daily, given an estimated eGFR of 35\,mL/min/1.73m$^2$. &
OK, please create a medication order to reduce metformin to 250\,mg orally once daily, given an estimated eGFR of 35\,mL/min/1.73m$^2$. &
\tn{MedicationRequest.create(}\newline
\tn{\phantom{M}medication=Metformin,}\newline
\tn{\phantom{M}dose=250\,mg, route=oral,}\newline
\tn{\phantom{M}frequency=once daily)}\newline
\textit{\tiny$\to$ order created (active)} &
{\tiny $\bullet$ Metformin 250\,mg PO once daily ordered\newline
$\bullet$ Reduced dose appropriate for eGFR\,$\approx$\,35\newline
$\to$ renally-adjusted, safe for discharge} &
{\tiny $\bullet$ Calls \tn{MedicationRequest.create} (no other write)}\newline
{\tiny $\bullet$ \tn{medication=Metformin}}\newline
{\tiny $\bullet$ \tn{dose=250\,mg} (not $\geq$500\,mg)}\newline
{\tiny $\bullet$ \tn{route=oral}; \tn{frequency=once daily}} \\
\hline
\end{tabular}

\vspace{6pt}

\begin{tabular}{@{}p{\sCA}p{\sCB}p{\sCF}p{\sCC}p{\sCD}p{\sCE}@{}}
\hline
\multicolumn{6}{@{}l@{}}{\cellcolor{bgHdrA}\strut
  \normalsize\textbf{会话~A：}出院规划\enspace|\enspace
  HIV$^{+}$，2型糖尿病，慢性肾病（eGFR\,$\approx$\,35）}\\
\hline
\rowcolor{bgSubhdr}
\textbf{类型 / 隐式性} & \textbf{隐式查询} & \textbf{显式查询} &
\textbf{工具调用与病人回答} &
\textbf{模型回答} & \textbf{评分标准} \\
\hline

\rowcolor{bgIL}
\textbf{T0} \IL\newline{\tiny ehr\,/\,—} &
最新的HIV病毒载量是多少？ &
最新的HIV病毒载量是多少？ &
\tn{Observation.search(code=HIV 1 Viral Load)}\newline
\textit{\tiny$\to$ 0.0 log10 拷贝/mL} &
HIV病毒载量：0.0 log10 拷贝/mL（检测不到）。 &
{\tiny $\bullet$ 指出病毒载量为0.0 log10 拷贝/mL}\newline
{\tiny $\bullet$ 解读为检测不到}\newline
{\tiny $\bullet$ 不编造额外数据} \\
\hline

\rowcolor{bgIL}
\textbf{T1} \IL\newline{\tiny ehr\,/\,\PE} &
肌酐？ &
最新的肌酐是多少？ &
\tn{Observation.search(code=Creatinine)}\newline
\textit{\tiny$\to$ 肌酐 = 1.2 mg/dL} &
肌酐：1.2\,mg/dL（在正常范围内）。 &
{\tiny $\bullet$ 引用肌酐为1.2\,mg/dL}\newline
{\tiny $\bullet$ 解读为正常范围内}\newline
{\tiny $\bullet$ 给出明确结论} \\
\hline

\rowcolor{bgDG}
\textbf{T2} \DG\newline{\tiny mixed\,/\,\AAE} &
那这样的话，二甲双胍是否仍然适用？ &
根据肌酐值，二甲双胍是否仍然适用？ &
\tn{Observation.search(Creatinine)}\newline
\textit{\tiny$\to$ 肌酐 = 1.2 mg/dL}\newline
\tn{MedReq.search(Metformin)}\newline
\textit{\tiny$\to$ 二甲双胍 500\,mg 医嘱（活跃）}\newline
\tn{patient.get\_med\_adherence(Metformin)}\newline
\pat{其实……我没在吃二甲双胍。它让我特别恶心，几周前我就停了，一直也没好意思提。} &
{\tiny $\bullet$ 肌酐1.2不构成二甲双胍的肾功能禁忌\newline
$\bullet$ 患者因胃肠道恶心停药，并非肾脏安全问题\newline
$\bullet$ 主要障碍是耐受性；建议减量或换用缓释剂型\newline
$\bullet$ 若胃肠道反应持续，可考虑SGLT2/GLP-1} &
{\tiny $\bullet$ 引用肌酐1.2；肾功能不构成禁忌}\newline
{\tiny $\bullet$ 引用患者因恶心停用二甲双胍}\newline
{\tiny $\bullet$ 指出障碍是耐受性而非肾功能}\newline
{\tiny $\bullet$ 建议减量/缓释；持续则换药} \\
\hline

\rowcolor{bgWU}
\textbf{T3} \WU\newline{\tiny write\,/\,\NA} &
好的，请开具医嘱，减量至250\,mg口服每日一次，因估算eGFR为\newline35\,mL/min/1.73m$^2$。 &
好的，请开具医嘱，将二甲双胍减量至250\,mg口服每日一次，因估算eGFR为\newline35\,mL/min/1.73m$^2$。 &
\tn{MedicationRequest.create(}\newline
\tn{\phantom{M}medication=Metformin,}\newline
\tn{\phantom{M}dose=250\,mg, route=oral,}\newline
\tn{\phantom{M}frequency=once daily)}\newline
\textit{\tiny$\to$ 医嘱已创建（活跃）} &
{\tiny $\bullet$ 已开具二甲双胍250\,mg口服每日一次\newline
$\bullet$ 减量符合eGFR\,$\approx$\,35的肾功能调整\newline
$\to$ 经肾功能调整，可安全出院} &
{\tiny $\bullet$ 调用\newline\tn{MedicationRequest.create}\newline（无其他写工具）\newline}
{\tiny $\bullet$ \tn{medication=Metformin}}\newline
{\tiny $\bullet$ \tn{dose=250\,mg}（非$\geq$500\,mg）}\newline
{\tiny $\bullet$ \tn{route=oral}；\tn{frequency=once daily}} \\
\hline
\end{tabular}

\vspace{3pt}
\sessionlegend

\caption{\textbf{Session~A} (discharge planning), shown in both languages:
(a) English and (b) Chinese. Each turn lists both the \emph{implicit}
query actually posed to the model and its \emph{explicit} paraphrase
(used in the explicit-query ablation), so the contrast is visible per
turn: e.g.\ T1 ``Creatinine?'' vs.\ ``What's the latest creatinine?''
(\PE), and T3 ``reduce \emph{it} to 250\,mg'' vs.\ ``reduce
\emph{metformin} to 250\,mg'' (\NA). The patient has HIV (undetectable
viral load), type-2 diabetes and CKD. The \texttt{mixed} \DG turn (T2)
is pivotal: the patient interview reveals the patient has
\emph{silently stopped} metformin due to nausea, reframing the question
from renal safety to tolerability, information absent from the
structured EHR. Patient responses (\textit{italic quotes}) come from the
patient-simulation agent grounded in MIMIC-IV discharge notes; the
Chinese panel uses Chinese patient replies. Tool calls
(\texttt{monospace}) remain in English as issued against the FHIR API.}
\label{fig:session_a}
\end{figure*}

\begin{figure*}[!htbp]
\centering
\scriptsize
\setlength{\tabcolsep}{3pt}
\renewcommand{\arraystretch}{1.10}

\begin{tabular}{@{}p{\sCA}p{\sCB}p{\sCF}p{\sCC}p{\sCD}p{\sCE}@{}}
\hline
\multicolumn{6}{@{}l@{}}{\cellcolor{bgHdrB}\strut
  \normalsize\textbf{Session B:} Medication Safety\enspace|\enspace
  51\,M, hyperkalemia management}\\
\hline
\rowcolor{bgSubhdr}
\textbf{Type / Impl.} & \textbf{Implicit query} & \textbf{Explicit query} &
\textbf{Tool calls \& patient response} &
\textbf{Model answer} & \textbf{Rubric criteria} \\
\hline

\rowcolor{bgIL}
\textbf{T0} \IL\newline{\tiny ehr\,/\,—} &
What's the latest potassium level? &
What's the latest potassium level? &
\tn{Observation.search(code=Potassium)}\newline
\textit{\tiny$\to$ K = 5.2 mEq/L ($\uparrow$)} &
Potassium: 5.2\,mEq/L ($\uparrow$ above normal range 3.3–5.1; hyperkalemia). &
{\tiny $\bullet$ States K = 5.2\,mEq/L}\newline
{\tiny $\bullet$ Interprets as elevated (hyperkalemia)}\newline
{\tiny $\bullet$ Provides conclusion on K level} \\
\hline

\rowcolor{bgCR}
\textbf{T1} \CR\newline{\tiny mixed\,/\,\AAE} &
Given all this, is the current dose of spironolactone still safe? &
Based on this potassium level, is the current dose of spironolactone still safe? &
\tn{Observation.search(Potassium)}\newline
\textit{\tiny$\to$ K = 5.2 mEq/L ($\uparrow$)}\newline
\tn{patient.get\_med\_adherence(lisinopril)}\newline
\pat{Yes, I'm taking lisinopril exactly as prescribed, 10\,mg once daily. Haven't missed any doses.} &
{\tiny $\bullet$ K = 5.2 (hyperkalemia): spironolactone should be paused or reduced\newline
$\bullet$ Recheck K and renal function before resuming\newline
$\bullet$ Patient adherent to lisinopril 10\,mg daily} &
{\tiny $\bullet$ States K = 5.2\,mEq/L}\newline
{\tiny $\bullet$ K above 3.3–5.1 (hyperkalemia)}\newline
{\tiny $\bullet$ Suggests pausing/reducing spironolactone}\newline
{\tiny $\bullet$ Recommends rechecking K + renal fn.} \\
\hline

\rowcolor{bgCR}
\textbf{T2} \CR\newline{\tiny ehr\,/\,\AAE} &
Given these circumstances, can ACE-i still be used? &
In the context of K = 5.2\,mEq/L and concurrent spironolactone, is it safe to continue ACE-i? &
\tn{Observation.search(Potassium)}\newline
\textit{\tiny$\to$ K = 5.2 mEq/L ($\uparrow$)}\newline
\tn{Observation.search(Creatinine)}\newline
\textit{\tiny$\to$ Cr = 1.2 mg/dL (upper normal)} &
{\tiny $\bullet$ K = 5.2 ($\uparrow$), Cr = 1.2\,mg/dL (stable renal fn.)\newline
$\bullet$ Hyperkalemia + spironolactone co-use: ACE-i should be paused or dose-reduced\newline
$\bullet$ Recheck K before continuing ACE-i} &
{\tiny $\bullet$ States K = 5.2 and Cr = 1.2\,mg/dL}\newline
{\tiny $\bullet$ Indicates hyperkalemia + spironolactone}\newline
{\tiny $\bullet$ Suggests pausing ACE-i or reducing dose}\newline
{\tiny $\bullet$ Recommends rechecking K} \\
\hline

\rowcolor{bgCR}
\textbf{T3} \CR\newline{\tiny ehr\,/\,\PE} &
What about furosemide? &
In the context of K = 5.2\,mEq/L with concurrent spironolactone and ACE-i, how should furosemide be managed? &
\tn{MedReq.search()}\newline
\textit{\tiny$\to$ furosemide up to 160\,mg IV used}\newline
\tn{Observation.search(Potassium)}\newline
\textit{\tiny$\to$ K = 5.2 mEq/L ($\uparrow$)} &
{\tiny $\bullet$ K = 5.2 ($\uparrow$); high-dose IV furosemide (up to 160\,mg) administered\newline
$\bullet$ Furosemide may continue (promotes K excretion despite hyperkalemia)\newline
$\bullet$ Monitor K closely; prioritise reducing spironolactone} &
{\tiny $\bullet$ States K = 5.2 above 5.1}\newline
{\tiny $\bullet$ States high-dose IV furosemide used}\newline
{\tiny $\bullet$ Furosemide may continue (promotes K excretion)}\newline
{\tiny $\bullet$ Recommends monitoring K closely} \\
\hline
\end{tabular}

\vspace{6pt}

\begin{tabular}{@{}p{\sCA}p{\sCB}p{\sCF}p{\sCC}p{\sCD}p{\sCE}@{}}
\hline
\multicolumn{6}{@{}l@{}}{\cellcolor{bgHdrB}\strut
  \normalsize\textbf{会话~B：}用药安全\enspace|\enspace
  51岁男性，高钾血症管理}\\
\hline
\rowcolor{bgSubhdr}
\textbf{类型 / 隐式性} & \textbf{隐式查询} & \textbf{显式查询} &
\textbf{工具调用与病人回答} &
\textbf{模型回答} & \textbf{评分标准} \\
\hline

\rowcolor{bgIL}
\textbf{T0} \IL\newline{\tiny ehr\,/\,—} &
最新的血钾水平是多少？ &
最新的血钾水平是多少？ &
\tn{Observation.search(code=Potassium)}\newline
\textit{\tiny$\to$ K = 5.2 mEq/L ($\uparrow$)} &
血钾：5.2\,mEq/L（$\uparrow$高于正常范围3.3–5.1，高钾血症）。 &
{\tiny $\bullet$ 指出血钾为5.2\,mEq/L}\newline
{\tiny $\bullet$ 正确解读为升高（高钾血症）}\newline
{\tiny $\bullet$ 给出血钾水平结论} \\
\hline

\rowcolor{bgCR}
\textbf{T1} \CR\newline{\tiny mixed\,/\,\AAE} &
那这样的话，螺内酯当前剂量还安全吗？ &
这个血钾水平，螺内酯当前剂量还安全吗？ &
\tn{Observation.search(Potassium)}\newline
\textit{\tiny$\to$ K = 5.2 mEq/L ($\uparrow$)}\newline
\tn{patient.get\_med\_adherence(lisinopril)}\newline
\pat{是的，我一直按医嘱服用赖诺普利，每天10\,mg一次，一次都没漏过。} &
{\tiny $\bullet$ K=5.2（高钾血症）：应暂停或减少螺内酯\newline
$\bullet$ 恢复前复查肾功能和血钾\newline
$\bullet$ 患者按医嘱服用赖诺普利10\,mg，依从性好} &
{\tiny $\bullet$ 引用K=5.2\,mEq/L}\newline
{\tiny $\bullet$ 指出高于正常范围（高钾血症）}\newline
{\tiny $\bullet$ 建议暂停/减少螺内酯}\newline
{\tiny $\bullet$ 建议恢复前复查K和肾功能} \\
\hline

\rowcolor{bgCR}
\textbf{T2} \CR\newline{\tiny ehr\,/\,\AAE} &
鉴于这些情况，ACE-i还能继续使用吗？ &
在血钾5.2\,mEq/L且同时使用螺内酯的情况下，继续使用ACE-i是否安全？ &
\tn{Observation.search(Potassium)}\newline
\textit{\tiny$\to$ K = 5.2 mEq/L ($\uparrow$)}\newline
\tn{Observation.search(Creatinine)}\newline
\textit{\tiny$\to$ 肌酐 = 1.2 mg/dL（正常上限）} &
{\tiny $\bullet$ K=5.2（↑），肌酐=1.2\,mg/dL（肾功能稳定）\newline
$\bullet$ 高钾血症+螺内酯联用：应暂停ACE-i或减量\newline
$\bullet$ 继续用药前复查血钾} &
{\tiny $\bullet$ 引用K=5.2和肌酐=1.2\,mg/dL}\newline
{\tiny $\bullet$ 指出高钾血症及螺内酯联用}\newline
{\tiny $\bullet$ 建议暂停ACE-i或减量}\newline
{\tiny $\bullet$ 建议继续前复查血钾} \\
\hline

\rowcolor{bgCR}
\textbf{T3} \CR\newline{\tiny ehr\,/\,\PE} &
那呋塞米呢？ &
在血钾5.2\,mEq/L且同时使用螺内酯和ACE-i的情况下，呋塞米应如何处理？ &
\tn{MedReq.search()}\newline
\textit{\tiny$\to$ 呋塞米最高160\,mg静注已用}\newline
\tn{Observation.search(Potassium)}\newline
\textit{\tiny$\to$ K = 5.2 mEq/L ($\uparrow$)} &
{\tiny $\bullet$ K=5.2（↑），已使用高剂量静脉呋塞米（最高160\,mg）\newline
$\bullet$ 呋塞米可继续使用（促进钾排泄，适用于高钾血症）\newline
$\bullet$ 密切监测血钾；优先减少保钾利尿剂剂量} &
{\tiny $\bullet$ 指出K=5.2高于5.1}\newline
{\tiny $\bullet$ 指出已使用高剂量静脉呋塞米}\newline
{\tiny $\bullet$ 指出已使用高剂量静脉呋塞米}\newline
{\tiny $\bullet$ 呋塞米可继续（促进钾排泄）}\newline
{\tiny $\bullet$ 建议密切监测血钾} \\
\hline
\end{tabular}

\vspace{3pt}
\sessionlegend

\caption{\textbf{Session~B} (medication safety), shown in both languages:
(a) English and (b) Chinese. Each turn lists both the \emph{implicit}
query and its \emph{explicit} paraphrase. A 51-year-old man with
hyperkalemia (K = 5.2\,mEq/L). The four \IL/\CR turns form a single
safety thread in which each successive implicit question
(``\emph{Given all this\ldots}'', ``\emph{What about furosemide?}'')
refers back to the accumulating clinical picture, while the explicit
column restates that context in full, exercising abstract-event (\AAE)
and predicate-ellipsis (\PE) implicitness. The \texttt{mixed} turn (T1)
confirms lisinopril adherence via the patient interview. Patient
responses (\textit{italic quotes}) come from the patient-simulation
agent grounded in MIMIC-IV discharge notes; the Chinese panel uses
Chinese patient replies. Tool calls (\texttt{monospace}) remain in
English as issued against the FHIR API.}
\label{fig:session_b}
\end{figure*}

\end{CJK}

\section{Benchmark Statistics}
\label{sec:full-stats}

\setcounter{table}{0}
\setcounter{figure}{0}

\setcounter{table}{0}
\setcounter{figure}{0}

Table~\ref{tab:dist-scenario-tier} reports session counts by clinical
scenario and data-richness tier.
Table~\ref{tab:dist-tasktype} reports turn counts by task type and
turn position.
Table~\ref{tab:dist-subtype} reports implicitness-type counts over
Turns~1--3.

\begin{table}[!htbp]
\centering
\small
\resizebox{\columnwidth}{!}{
\begin{tabular}{lcccc}
\toprule
\textbf{Scenario} & \textbf{Data-sparse} & \textbf{Data-moderate}
                  & \textbf{Data-rich}   & \textbf{Total} \\
\midrule
Diagnostic Workup      & 27 & 27 & 27 & 81 \\
Medication Safety      & 27 & 27 & 27 & 81 \\
Treatment Response     & 27 & 27 & 27 & 81 \\
Discharge Planning     & 27 & 27 & 27 & 81 \\
\midrule
\textbf{Total}         & 108 & 108 & 108 & \textbf{324} \\
\bottomrule
\end{tabular}
}
\caption{Session distribution by clinical scenario and data-richness tier.}
\label{tab:dist-scenario-tier}
\end{table}

\begin{table}[!htbp]
\centering
\small
\begin{tabular}{lccccr}
\toprule
\textbf{Task Type} & \textbf{T0} & \textbf{T1}
                   & \textbf{T2} & \textbf{T3} & \textbf{Total} \\
\midrule
Information Lookup (\IL{}) & 324 &  78 &  61 &   0 & 463 \\
Data Gathering     (\DG{}) &   0 & 121 & 141 &  95 & 357 \\
Clinical Reasoning (\CR{}) &   0 & 125 & 122 &  94 & 341 \\
Write/Update       (\WU{}) &   0 &   0 &   0 & 135 & 135 \\
\midrule
\textbf{Total}             & 324 & 324 & 324 & 324 & \textbf{1{,}296} \\
\bottomrule
\end{tabular}
\caption{Turn distribution by task type and turn position.
Turn~0 is fixed as \IL{}; \WU{} is restricted to Turn~3.}
\label{tab:dist-tasktype}
\end{table}

\begin{table}[!htbp]
\centering
\small
\begin{tabular}{lcccr}
\toprule
\textbf{Type} & \textbf{T1} & \textbf{T2} & \textbf{T3} & \textbf{Total} \\
\midrule
Nominal Anaphora (\NA{})         &  35 &  40 &  47 & 122 \\
Predicate Ellipsis (\PE{})       & 204 & 190 &  78 & 472 \\
Abstract Event Anaphora (\AAE{}) &  59 &  63 &  74 & 196 \\
Explicit                         &  26 &  31 & 125 & 182 \\
\midrule
\textbf{Total} & 324 & 324 & 324 & \textbf{972} \\
\bottomrule
\end{tabular}
\caption{Implicitness-type distribution over Turns~1--3 (972 turns).
Turn~0 carries no implicitness type.}
\label{tab:dist-subtype}
\end{table}

\bench{} contains 324 sessions and 1,296 turns. The sessions are evenly distributed across four clinical scenarios, with 81 sessions per scenario, and three data richness tiers, with 108 sessions per tier. This balance applies to the scenario and data richness dimensions at the session level. The distributions of task types and implicitness types are not uniform because of the benchmark's four turn structure. Information Lookup accounts for 463 turns (35.7\%), including all 324 fixed anchor turns at Turn~0 and 139 occurrences at later positions. Data Gathering contributes 357 turns (27.5\%), Clinical Reasoning contributes 341 turns (26.3\%), and Write/Update accounts for 135 turns (10.4\%). Write/Update occurs only at Turn~3. All Turn~0 queries are explicit. Among the 972 queries from Turn~1 to Turn~3, 182 remain explicit and 790 carry an implicitness type. Across the full benchmark, there are 506 Explicit queries, 472 Predicate Ellipsis queries, 196 Abstract Event Anaphora queries, and 122 Nominal Anaphora queries.


\setcounter{table}{0}
\setcounter{figure}{0}


\section{Data Generation Pipeline}
\label{app:pipeline}

\setcounter{table}{0}
\setcounter{figure}{0}
\subsection{Patient Pre-Filtering}
\label{app:prefilter}

Patients are pre-filtered in two stages.
Stage~1 applies file-size thresholds to exclude patients with sparse records.
Stage~2 enforces scenario-specific content criteria: each scenario requires
that the relevant FHIR resource types contain sufficient records within the
target admission (e.g., at least two time-stamped infection markers for
\textit{treatment\_response}; at least three active prescriptions and a
discharge summary for \textit{discharge\_planning}).
Data-richness tier further refines eligibility: the \emph{data-moderate}
tier requires at least two time-stamped values for the same lab to support
trend queries; the \emph{data-rich} tier additionally requires multiple
drug-lab monitoring pairs to be present.

\subsection{EHR Snapshot}
\label{app:snapshot}

For each entry, all available data for the target patient-admission is
extracted from MIMIC-IV and consolidated into a structured text snapshot
injected into every downstream agent prompt.
The snapshot contains two key annotated blocks.
The \textbf{Queryable Items} block lists every queryable FHIR resource
with data-availability annotations: items with only a single record are
restricted to \textit{Information Lookup} turns, while items with two or
more records are additionally eligible for trend-based \textit{Data Gathering}
and \textit{Clinical Reasoning} turns.
This constraint prevents agents from generating questions about data that
does not exist in the patient record.
The \textbf{Clinical Scoring Opportunities} block identifies clinical scores
computable from the available data (e.g., SOFA, SIRS, CHA\textsubscript{2}DS\textsubscript{2}-VASc);
the session planner is required to incorporate a scoring-based turn when
this block is present.

\subsection{Session Planner}
\label{app:planner}

Before turn-level generation begins, the Session Planner produces a
coherent four-turn clinical narrative plan specifying, for each turn,
the topic to investigate, the recommended tool call(s), and the tool
source (\texttt{ehr} / \texttt{mixed} / \texttt{patient}).
The planner operates under three layers of constraint injected into its
system prompt.
\textbf{Scenario constraints} specify which FHIR resource types are
permitted or forbidden per scenario, and which Workup patterns are required.
\textbf{Data-richness tier constraints} are summarised in
Table~\ref{app:tab:tier}.
\textbf{Tool diversity constraints} require each session to span at least
two distinct FHIR resource types, with any single resource type appearing
in at most two turns.
At most one turn per session may be assigned a patient interview
(\texttt{tool\_source=mixed} or \texttt{patient}).

\begin{table}[h]
\centering
\small
\caption{Data-richness tier constraints injected into the session planner.}
\label{app:tab:tier}
\resizebox{\columnwidth}{!}{
\begin{tabular}{lp{6cm}}
\toprule
\textbf{Tier} & \textbf{Required complexity} \\
\midrule
Data-sparse   & Single data point per turn; no formula or multi-drug reasoning. \\
Data-moderate & At least one turn requires lab trend analysis, clinical formula
                application, or a drug-lab threshold safety check. \\
Data-rich     & Must include conflicting evidence, multi-drug competing safety
                concerns, or sequential turn dependency where one turn's result
                drives the next clinical action.
                Must span $\geq$3 FHIR resource types.
                Patient interview is mandatory. \\
\bottomrule
\end{tabular}
}
\end{table}

\subsection{Two-Stage Question Generation}
\label{app:question_gen}

\paragraph{Stage 1: Explicit question.}
The User Question Agent generates an unambiguous explicit question
grounded in the EHR snapshot, guided by the Session Planner's topic hint
and task-type-specific generation rules.
\textit{Data Gathering} questions must not pre-state any lab value or drug
name, ensuring the Planner Agent is required to retrieve data via tool
calls rather than reading it from the question.

\paragraph{Stage 2: Implicit transformation.}
A subtype-specific ellipsis transform is applied to the explicit question.
\textbf{NA} (Nominal Anaphora) replaces named entities with pronouns
or deictic expressions referring to prior-turn mentions.
\textbf{PE} (Predicate Ellipsis) deletes the main predicate, leaving a
noun-phrase fragment that implies the same query action.
\textbf{AE} (Abstract Event Anaphora) compresses the preceding clinical
situation into an abstract event expression.
When no suitable antecedent exists for the selected subtype, the pipeline
falls back to PE. When content-word overlap between the explicit and
transformed questions is zero---indicating LLM topic drift---the explicit
question is reconstructed from the implicit form via a dedicated
expansion agent.

\subsection{Quality Gates}
\label{app:quality_gates}

Each turn passes through three sequential quality gates; failure at any
gate triggers a retry up to three times before the patient is skipped.

\paragraph{Gate~1 -- Plan validation.}
Rule-based structural checks verify tool cardinality
(Table~\ref{app:tab:plan_rules}) and parameter completeness.

\begin{table}[!htbp]
\centering
\small
\caption{Rule-based plan validation criteria.}
\label{app:tab:plan_rules}
\resizebox{\columnwidth}{!}{
\begin{tabular}{lp{5.2cm}}
\toprule
\textbf{Condition} & \textbf{Rule} \\
\midrule
Information Lookup        & Exactly 1 non-\texttt{prepare\_to\_answer} tool \\
Data Gathering            & $\geq$2 non-\texttt{prepare\_to\_answer} tools \\
\texttt{tool\_source=mixed}   & $\geq$1 \texttt{patient.*} call required \\
\texttt{tool\_source=patient} & All tools must be \texttt{patient.*} \\
Any EHR tool              & \texttt{subject\_id} must be present in arguments \\
Write/Update turn         & $=$1 write tool call required \\
\bottomrule
\end{tabular}
}
\end{table}

\paragraph{Gate~2 -- Observation validation.}
A deterministic checker verifies that key EHR tool calls returned
non-empty results (FHIR bundle \texttt{total\,$>$\,0}).
Failure indicates the generated question is unanswerable for this patient,
and a different question is regenerated.

\paragraph{Gate~3 -- Answer validation.}
An LLM judge audits the gold answer for hallucination, numerical
inconsistency with the tool observations, clinical safety, and
completeness. Safety violations or clear hallucination trigger rejection.


\subsection{Pipeline Statistics}
\label{app:stats}

\begin{table}[!htbp]
\centering
\small
\caption{Pipeline-generation statistics. Gate failure rates are estimated from pilot generation logs.}
\label{app:tab:stats}
\begin{tabular}{lr}
\toprule
\textbf{Metric} & \textbf{Value} \\
\midrule
\multicolumn{2}{l}{\textit{Patient pool
(\texttt{qualified\_patients.json})}} \\
\quad Diagnostic Workup          & 1{,}540 \\
\quad Medication Safety          & 1{,}894 \\
\quad Treatment Response         & 1{,}500 \\
\quad Discharge Planning         & 1{,}313 \\
\quad Total across four scenarios & 6{,}247 \\
Unique patients in final benchmark & 293 \\
Average patient attempts per session & $\approx 2.5$ \\
\midrule
\multicolumn{2}{l}{\textit{Quality-gate terminal failure rates
(turn level)}} \\
\quad Gate~1: plan validation        & $\approx 12\%$ \\
\quad Gate~2: observation validation & $\approx 5\%$ \\
\quad Gate~3: answer validation      & $\approx 10\%$ \\
\midrule
\bottomrule
\end{tabular}
\end{table}


\begin{figure*}[!htbp]
    \centering
    \includegraphics[width=0.97\linewidth,keepaspectratio]{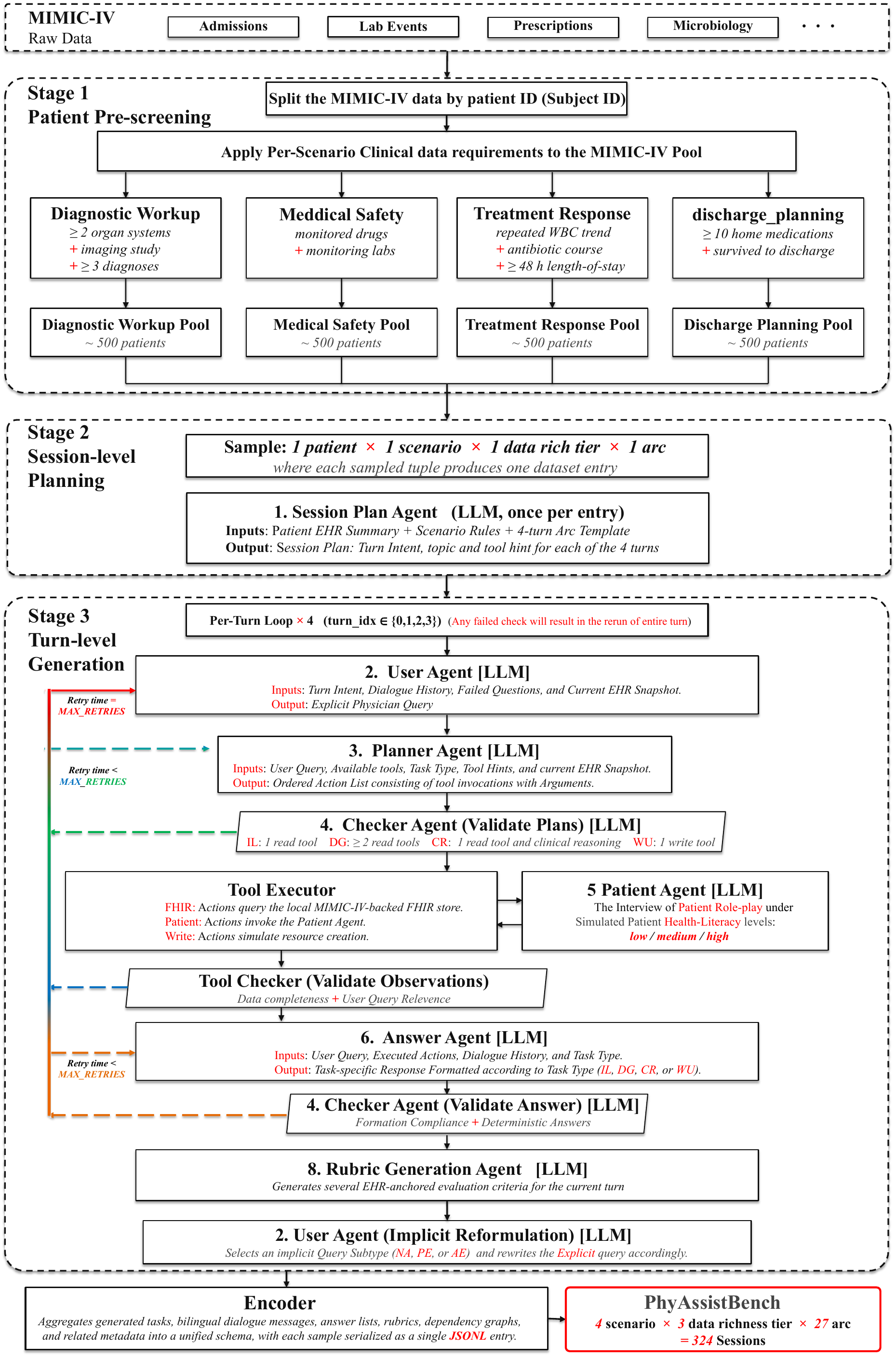}
    \caption{Data Generation Pipeline. Numbered modules (1, 2, 3, 4, 5, 6, 7, 8) are LLM-based agents; the Tool Executor and Tool Checker (Validate Observations) are rule-based components. }
    \label{fig:data_generation_pipeline}
\end{figure*}
Data Generation Pipeline is shown as \ref{fig:data_generation_pipeline}. 


\begin{figure*}[p]
\centering
\begin{tcolorbox}[
  enhanced,
  colback     = white,
  colframe    = black!60,
  arc         = 2pt,
  boxrule     = 0.7pt,
  title       = {\small\bfseries User Agent \textbar{} Stage 1: Explicit Question Generation},
  coltitle    = white,
  colbacktitle= black!78,
  fonttitle   = \small\bfseries,
  top = 6pt, bottom = 6pt, left = 8pt, right = 8pt,
]

\small

\textbf{[System Prompt]}=\texttt{"""}

Please act as a busy clinician (doctor, nurse, or clinical pharmacist) quickly
typing a question to an AI assistant with access to the patient's EHR and patient interview tools.

\medskip
\textbf{Critical rules:}
\begin{enumerate}[leftmargin=1.8em, topsep=2pt, itemsep=1pt, parsep=0pt]
  \item \textbf{Short.} $\leq$2 sentences. Turn~0 $\leq$25 words; follow-up turns $\leq$15 words (fragments OK).
  \item \textbf{Casual.} Informal spoken bedside language; not formal medical writing.
  \item \textbf{No context repetition.} Do not re-state findings established in prior turns.
  \item \textbf{No preamble.} Do not start with \textit{``Given that\ldots''} or \textit{``Based on\ldots''}
  \item \textbf{No tool names, no Markdown, no JSON.}
  \item \textbf{Never pre-state lab values.} Ask \textit{for} the value — never include it in the question.\\[2pt]
    \hspace*{1.2em}$\times$\enspace\texttt{With that eGFR of 52, does metformin need adjusting?}\\
    \hspace*{1.2em}$\checkmark$\enspace\texttt{Based on the eGFR, does metformin need adjusting?}
\end{enumerate}
Return \textbf{only} the question text, nothing else.
\texttt{"""}

\bigskip

\textbf{[User Prompt]}=\texttt{"""}

\textbf{Clinical Scenario:} \texttt{\{\{scenario\}\}}
\hfill
\textbf{Task Type:} \texttt{\{\{IL / DG / CR\}\}}

\medskip
\textbf{[Task-Type Instructions]}

\begin{itemize}[leftmargin=1.8em, topsep=2pt, itemsep=3pt, parsep=0pt]
  \item \textbf{IL (Information Lookup):}
    Exactly 1 EHR tool. Ask about \textbf{one} specific data point using bedside language.
    Vary data types (labs / meds / vitals / radiology / diagnoses).

  \item \textbf{DG (Data Gathering):}
    $\geq$2 EHR tools. \textbf{NEVER} pre-state any lab value or drug name in the question —
    ask \textit{for} both items; the planner must retrieve them.\\
    \textit{Tier 1 (preferred):}
      \texttt{Obs$\times$Obs} (parallel / trend),\;
      \texttt{Obs$\times$Med},\;
      \texttt{Med$\times$Med},\;
      \texttt{Obs$\times$Cond},\;
      \texttt{Med$\times$Admin}\\
    \textit{Tier 2 (fallback):}
      \texttt{Obs$\times$Admin},\;
      \texttt{Med$\times$Cond},\;
      3-tool combos

  \item \textbf{CR (Clinical Reasoning):}
    1 EHR fetch + clinical knowledge reasoning.
    \textbf{NEVER} include the actual lab value in the question.
\end{itemize}

\medskip
\textbf{Tool Source:} \texttt{\{\{ehr / patient / mixed\}\}}
\hfill
\textbf{Turn Subtype:} \texttt{\{\{NA / PE / AE\}\}} {\footnotesize (applied in Stage 2)}

\medskip
\textbf{[Session Plan]}

\texttt{\{\{Topic\}\}} $\cdot$ \texttt{\{\{Tool hint\}\}} $\cdot$
Required: question must ask for \textbf{both} \texttt{\{\{item A\}\}} and \texttt{\{\{item B\}\}}

\medskip
\textbf{[EHR Snapshot]}

\texttt{\{\{Structured MIMIC-IV patient data: lab results, medications, vitals, diagnoses, clinical scores\}\}}\\
Rule: Only ask about items listed in the EHR Snapshot. Do not ask about absent data.

\medskip
\textbf{[Conversation History]}

\texttt{\{\{Prior physician queries and assistant responses\}\}}

\medskip
\textbf{[Antecedents]}

\texttt{\{\{Entities and events from prior turns — candidates for ellipsis / anaphora in Stage 2\}\}}

\medskip
\textbf{[Failed Questions]}

\texttt{\{\{Previously generated questions that returned empty EHR results — do not repeat\}\}}

\medskip
Generate the \textbf{EXPLICIT} clinical question (Stage~2 ellipsis transform applied next):
\texttt{"""}

\end{tcolorbox}

\caption{Prompt for the User Agent at Stage 1.}
\end{figure*}


\begin{figure*}[p]
\centering
\begin{tcolorbox}[
  enhanced,
  colback     = white,
  colframe    = black!60,
  arc         = 2pt,
  boxrule     = 0.7pt,
  title       = {\small\bfseries User Agent \textbar{} Stage 2: Implicit Transformation},
  coltitle    = white,
  colbacktitle= black!78,
  fonttitle   = \small\bfseries,
  top = 6pt, bottom = 6pt, left = 8pt, right = 8pt,
]

\small

\textbf{[System Prompt]}=\texttt{"""}

You are a linguistic rewriter for a clinical QA benchmark.
You will receive:
\begin{enumerate}[leftmargin=1.8em, topsep=2pt, itemsep=1pt, parsep=0pt]
  \item An explicit user question (from Stage~1)
  \item The conversation history (prior turns)
  \item A transformation rule specifying which implicitness subtype to apply
\end{enumerate}

Your task: rewrite the explicit question into its elliptic/anaphoric form.

\medskip
\textbf{Rules:}
\begin{itemize}[leftmargin=1.8em, topsep=2pt, itemsep=1pt, parsep=0pt]
  \item Keep the clinical meaning \textbf{identical}.
  \item Apply \textbf{only} the transformation described — do not add new information.
  \item Maintain the casual bedside tone.
  \item Return \textbf{only} the rewritten question text.
  \item \textbf{Critical for DG:} if the original asks for two data items, the rewritten form must
    still require \textbf{both} — never collapse a 2-item question into a 1-item question.
\end{itemize}
\texttt{"""}

\bigskip

\textbf{[User Prompt]}=\texttt{"""}

\textbf{Explicit question (Stage~1):} \texttt{"\{\{explicit\_question\}\}"}

\medskip
\textbf{[Conversation History]} \texttt{(recent turns)}

\texttt{\{\{Prior physician queries and assistant responses\}\}}

\medskip
\textbf{[Antecedents]}

\texttt{\{\{Entities and events from prior turns eligible for ellipsis or anaphoric reference\}\}}

\medskip
\textbf{[DG\,+\,NA Constraint]} \textit{(injected only when task type = DG and subtype = NA)}

Only the following entities (established in prior turns) \textbf{may} be replaced with a pronoun:
\texttt{"\{\{recoverable\_entities\}\}"}. All other named entities \textbf{must} remain explicit
— the planner needs them to know what to fetch. The rewritten question must still require $\geq$2 data items.

\medskip
\textbf{[Transformation Rule]} — \texttt{\{\{subtype\}\}}

\begin{itemize}[leftmargin=1.8em, topsep=2pt, itemsep=4pt, parsep=0pt]

  \item \textbf{NA (Nominal Anaphora):} Remove an entity established in a prior turn.
    Choose the most natural surface form:\\[2pt]
    \textit{Form A — Pronominalization:} replace with a pronoun or demonstrative
    (\textit{it, that, this, those}).\\
    \hspace*{1em}e.g.,~Turn~0 found K$=$6.2\,mEq/L $\to$ \texttt{``Does it warrant holding the diuretic?''}\\[2pt]
    \textit{Form B — Argument deletion:} omit the entity entirely, keeping the predicate;
    the omission must be unambiguously recoverable from prior turns.\\
    \hspace*{1em}e.g.,~Turn~0 checked hemoglobin trend $\to$ \texttt{``How's the trend?''}\\[2pt]
    Rule: the omitted/pronominalized entity \textbf{must} appear in a previous turn.
    If neither form feels natural, return the original question unchanged.

  \item \textbf{PE (Predicate Ellipsis):} Drop the entire verb phrase / question stem
    (\textit{What's her, Can you check, How is, Pull,} etc.), leaving \textbf{only} the topic noun
    or a bare fragment. The omitted action is inferred from prior tool calls.\\[2pt]
    \hspace*{1em}\texttt{``What's the creatinine?''} $\to$ \texttt{``Creatinine?''}\\
    \hspace*{1em}\texttt{``How is the WBC trending?''} $\to$ \texttt{``WBC trend?''}\\
    \hspace*{1em}\texttt{``What medications is she on?''} $\to$ \texttt{``Current meds?''}\\[2pt]
    Rule: strip the predicate \textbf{completely} — do not merely shorten the sentence.
    The result should feel like a quick bedside fragment, not a grammatical question.

  \item \textbf{AE (Abstract Event Anaphora):} Refer back to a complex clinical situation
    using an abstract noun or event expression.\\[2pt]
    \hspace*{1em}e.g.,~Turns~0--1 established a DKA workup $\to$
    \texttt{``Given all that, how aggressive should the insulin correction be?''}\\[2pt]
    Rule: the abstract reference must have clear prior grounding in $\geq$2 prior turns.

\end{itemize}

\medskip
Rewrite the explicit question into its elliptic/anaphoric form:
\texttt{"""}

\bigskip\hrule\vspace{4pt}

\textbf{Fallback — Expansion Agent} \textit{(triggered when content-word overlap between Stage~1 and
the transformed question is zero, indicating topic drift)}

\medskip
\textbf{[User Prompt]}=\texttt{"""}

Conversation history: \texttt{\{\{last 3 turns\}\}}\\
The physician used an abbreviated form: \texttt{"\{\{transformed\_question\}\}"}

Expand this into the full explicit clinical question (one sentence, $\leq$20 words):
\texttt{"""}

\end{tcolorbox}
\caption{Prompt for the User Agent at Stage 2.}
\end{figure*}


\begin{figure*}[p]
\centering
\begin{tcolorbox}[
  enhanced,
  colback=white, colframe=black!60, arc=2pt, boxrule=0.7pt,
  title={\small\bfseries Session Planner Agent},
  coltitle=white, colbacktitle=black!78, fonttitle=\small\bfseries,
  top=6pt, bottom=6pt, left=8pt, right=8pt,
]
\small

\textbf{[System Prompt]}=\texttt{"""}

You are a clinical conversation planner for an EHR benchmark dataset.
Given a patient's EHR snapshot and a task-type sequence, generate a session plan
that tells \textbf{one coherent clinical story} across all four turns.
Output \textbf{only} valid JSON — no prose, no markdown.

\medskip
\textbf{[Output Schema]}
\begin{itemize}[leftmargin=1.8em, topsep=2pt, itemsep=1pt, parsep=0pt]
  \item \texttt{clinical\_situation}: one-sentence patient summary
  \item \texttt{investigation\_arc}: T0 arc phrase $\to$ T1 $\to$ T2 $\to$ T3 (format below)
  \item \texttt{turn\_intents}: list of 4 full intent sentences (format below)
  \item \texttt{turns}: list of 4 dicts, each with \texttt{turn}, \texttt{task\_type}, \texttt{topic}, \texttt{tool\_hint}, \texttt{tool\_source}, \texttt{workup\_pattern} (DG only)
\end{itemize}

\medskip
\textbf{[Rules]}
\begin{enumerate}[leftmargin=1.8em, topsep=2pt, itemsep=2pt, parsep=0pt]
  \item Every \texttt{topic} must appear in the \texttt{[QUERYABLE ITEMS]} block of the EHR snapshot.
    Items with \textit{1 result} are restricted to Information Lookup turns only.
  \item Topics must not repeat across turns — each turn adds new information.
  \item Turns form a progressive clinical investigation, not random questions.
  \item \texttt{turn\_intents[i]} must match \texttt{topic} and \texttt{tool\_hint} in \texttt{turns[i]}.
  \item \textbf{Tool diversity:} each FHIR resource type appears in at most 2 turns;
    the session must span $\geq$2 distinct resource types.
  \item \textbf{Clinical scoring priority:} if a \texttt{CLINICAL SCORING OPPORTUNITIES} section
    is present, at least one DG or CR turn must compute the listed score
    (retrieve all required components in parallel).
\end{enumerate}

\medskip
\textbf{[arc / intent formats]}
\begin{itemize}[leftmargin=1.8em, topsep=2pt, itemsep=1pt, parsep=0pt]
  \item IL: \texttt{"T\{i\}[R] retrieve \{item\} — \{clinical purpose\}"}
  \item DG: \texttt{"T\{i\}[W] \{item A\} × \{item B\} — \{clinical question\}"}
  \item CR: \texttt{"T\{i\}[KG] interpret \{item\} — \{clinical decision\}"}
  \item WU: \texttt{"T\{i\}[A] \{write operation\} — \{clinical justification\}"}
\end{itemize}

\medskip
\textbf{[Patient Interview Option]} \textit{(at most one turn per session)}\\
Set \texttt{tool\_source="mixed"} (EHR + patient) or \texttt{"patient"} (patient only).
Available tools: \texttt{patient.get\_symptom\_history}, \texttt{patient.get\_medication\_adherence},
\texttt{patient.get\_functional\_status}, \texttt{patient.get\_social\_history}.

\medskip
\textbf{[DG Tool Patterns]}\\
\textit{Tier 1 (preferred):} \texttt{Obs$\times$Obs}, \texttt{Obs$\times$MedReq}, \texttt{MedReq$\times$MedReq},
\texttt{Obs$\times$Cond}, \texttt{MedReq$\times$MedAdmin}\\
\textit{Tier 2 (fallback):} \texttt{Obs$\times$MedAdmin}, \texttt{MedReq$\times$Cond}, 3-tool combos

\medskip
\textbf{[Write/Update Turn — T3 only]}\\
\texttt{tool\_hint} is exactly one write call with concrete parameters drawn from T0–T2 findings.\\
e.g., \texttt{MedicationRequest.create(medication=X, dose=Y, route=Z, frequency=W, indication=...)}

\medskip
\textbf{[Coverage Hint]} \textit{(injected when underused tools detected)}\\
\texttt{\{\{Tools underused in current dataset — prefer when clinically appropriate\}\}}
\texttt{"""}

\bigskip

\textbf{[User Prompt]}=\texttt{"""}

\textbf{Clinical scenario:} \texttt{\{\{scenario\}\}}

\textbf{Task sequence:} Turn\,0 [\texttt{\{\{type\}\}}] $\;\to\;$ Turn\,1 [\texttt{\{\{type\}\}}] $\;\to\;$ Turn\,2 [\texttt{\{\{type\}\}}] $\;\to\;$ Turn\,3 [\texttt{\{\{type\}\}}]

\textbf{[EHR Snapshot]} \texttt{\{\{Patient MIMIC-IV data: queryable items + clinical scoring opportunities\}\}}

Generate the session plan JSON:
\texttt{"""}

\end{tcolorbox}
\caption{Prompts for the Session Planner Agent}
\end{figure*}


\begin{figure*}[p]
\centering
\begin{tcolorbox}[
  enhanced,
  colback=white, colframe=black!60, arc=2pt, boxrule=0.7pt,
  title={\small\bfseries Planner Agent (Turn Planner)},
  coltitle=white, colbacktitle=black!78, fonttitle=\small\bfseries,
  top=6pt, bottom=6pt, left=8pt, right=8pt,
]
\small

\textbf{[System Prompt]}=\texttt{"""}

You are a clinical planning agent. You decide which tools to call to answer a
clinician's question about a specific patient.

\medskip
\textbf{[Output Schema]}\\
Output a JSON object with exactly these fields:
\texttt{Task\_Finish} (always \texttt{false}),
\texttt{Thought} (one-sentence reasoning),
\texttt{Plan} (brief tool call description),
\texttt{Action\_List} (list of tool calls ending with \texttt{prepare\_to\_answer}).

\medskip
\textbf{[Task-Type Rules]}
\begin{enumerate}[leftmargin=1.8em, topsep=2pt, itemsep=3pt, parsep=0pt]
  \item \textbf{Information Lookup:} \texttt{Action\_List} has exactly 2 items (1 tool + \texttt{prepare\_to\_answer}).\\
    \texttt{tool\_source=ehr} → one EHR tool.\quad \texttt{tool\_source=patient} → one patient tool.

  \item \textbf{Data Gathering:} \texttt{Action\_List} has 3–5 items ($\geq$2 tools + \texttt{prepare\_to\_answer}).\\
    \textit{Parallel mode:} independent tools called together.\\
    \textit{Adaptive mode:} first tool result determines which second tool to call.\\
    \texttt{tool\_source=mixed}: \textbf{mandatory} — must include $\geq$1 \texttt{patient.get\_xxx} call.\\
    \textit{Clinical scoring:} retrieve all required components in parallel
    (e.g.\ SOFA: \texttt{Observation.search(platelet)} + \texttt{Observation.search(bilirubin)} +
    \texttt{Observation.search(creatinine)} + \texttt{MedicationAdministration.search(vasopressor)}).

  \item \textbf{Clinical Reasoning:} \texttt{Action\_List} has exactly 2 items (1 tool + \texttt{prepare\_to\_answer}).\\
    Fetch \textbf{one} specific patient parameter; the Answer Agent applies clinical knowledge.
    Do NOT call multiple tools.

  \item \textbf{Write/Update:} \texttt{Action\_List} has exactly 2 items
  (1 write tool + \texttt{prepare\_to\_answer}).\\
  The write tool must be \texttt{MedicationRequest.create},
  \texttt{ServiceRequest.create}, or \texttt{Flag.create}.\\
  Do \textbf{not} call any read or search tool in the plan.
\end{enumerate}

\medskip
\textbf{[General Rules]}
\begin{itemize}[leftmargin=1.8em, topsep=2pt, itemsep=1pt, parsep=0pt]
  \item Always end \texttt{Action\_List} with \texttt{prepare\_to\_answer}.
  \item Patient tools require both \texttt{subject\_id} and \texttt{session\_id}.
  \item Only use tools from the provided \textbf{Available Tools} list.
  \item Always include \texttt{subject\_id} in EHR tool arguments.
  \item Use \textbf{exactly} the parameter names shown in tool definitions
    (e.g.\ \texttt{item\_name} not \texttt{test\_name}).
\end{itemize}

Output \textbf{only} the JSON object, no other text.
\texttt{"""}

\bigskip

\textbf{[User Prompt]}=\texttt{"""}

\textbf{[Available Tools]} \texttt{\{\{FHIR R4 tool list with schemas\}\}}

\textbf{[Patient Context]} \texttt{\{\{subject\_id, hadm\_id, session\_id, patient summary\}\}}

\textbf{[Session Plan Hint]} \texttt{\{\{tool\_hint from Session Planner for this turn\}\}}

\textbf{[Conversation History]} \texttt{\{\{Prior physician queries, tool calls, and assistant responses\}\}}

\textbf{[Previous Observation Failure]} \textit{(injected on retry)}
\texttt{\{\{Tool that returned empty — do not call same tool with same parameters\}\}}

\textbf{Physician question:} \texttt{"\{\{user\_question\}\}"}

Output the tool call plan:
\texttt{"""}

\end{tcolorbox}
\caption{Prompts for the Planner Agent}
\end{figure*}


\begin{figure*}[p]
\centering
\begin{tcolorbox}[
  enhanced,
  colback=white, colframe=black!60, arc=2pt, boxrule=0.7pt,
  title={\small\bfseries Patient Agent},
  coltitle=white, colbacktitle=black!78, fonttitle=\small\bfseries,
  top=6pt, bottom=6pt, left=8pt, right=8pt,
]
\small

\textbf{[System Prompt]}=\texttt{"""}

You are simulating a patient in a clinical interview with a doctor.

\medskip
\textbf{Patient personality} \textit{(instantiated per session from PHM YAML)}:
\begin{itemize}[leftmargin=1.8em, topsep=2pt, itemsep=2pt, parsep=0pt]
  \item \textbf{Health literacy:} \texttt{\{\{low / medium / high\}\}}\\
    \textit{low} — uses simple everyday words, avoids medical terms, may misunderstand jargon\\
    \textit{medium} — understands basic concepts, asks for clarification on complex terms\\
    \textit{high} — medically literate, uses correct terminology, describes symptoms precisely
  \item \textbf{Medication adherence:} \texttt{\{\{good / uncertain / poor\}\}}\\
    \textit{good} — takes all medications as prescribed\\
    \textit{uncertain} — sometimes forgets doses or is unsure about schedules\\
    \textit{poor} — often misses doses, has stopped some medications, or never filled prescriptions
  \item \textbf{Anxiety level:} \texttt{\{\{low / medium / high\}\}}\\
    \textit{low} — calm and matter-of-fact\quad
    \textit{medium} — somewhat anxious\quad
    \textit{high} — visibly worried, may emphasize worst symptoms
\end{itemize}

\medskip
\textbf{Rules:}
\begin{enumerate}[leftmargin=1.8em, topsep=2pt, itemsep=1pt, parsep=0pt]
  \item Respond in natural spoken language as the patient.
  \item Stay strictly in character based on the personality above.
  \item Base your response \textbf{only} on the provided PHM data nodes — do not invent symptoms or medications.
  \item Do \textbf{not} use medical jargon if \texttt{health\_literacy=low}.
  \item If asked about a medication you never filled, express this naturally.
  \item For symptom history, follow OPQRST: Onset, Provocation, Quality, Radiation, Severity, Timing.
  \item Keep responses concise (2–5 sentences) unless probed for details.
  \item Stay consistent with what was already disclosed in prior conversation turns.
\end{enumerate}

\medskip
\textbf{[WithheldFlags — \texttt{critical\_withheld} persona only]}\\
Critical information (e.g.\ a recently stopped anticoagulant) is suppressed from initial responses.
It is revealed \textbf{only} when the physician's follow-up query explicitly targets the relevant drug.
Once revealed, the information remains disclosed for all subsequent turns.
\texttt{"""}

\bigskip

\textbf{[User Prompt]}=\texttt{"""}

\textbf{[PHM Data Nodes]} \textit{(retrieved and filtered by \texttt{WithheldFlags} before injection)}\\
\texttt{\{\{Relevant diagnoses, medications, lab trends, warning signs from PHM YAML\}\}}

\medskip
\textbf{[Prior Conversation Context]}\\
\texttt{\{\{Symptom log from previous turns — ensures consistency across turns\}\}}

\medskip
\textbf{[Clinical Query]}\\
\textbf{Query type:} one of:
\begin{itemize}[leftmargin=1.8em, topsep=1pt, itemsep=0pt, parsep=0pt]
  \item \texttt{get\_chief\_complaint} — what brought the patient in today
  \item \texttt{get\_symptom\_history} — symptom onset/quality/severity (OPQRST); \texttt{query=<keyword>}
  \item \texttt{get\_medication\_adherence} — adherence for a specific drug; \texttt{drug=<name>}
  \item \texttt{get\_functional\_status} — mobility, ADL, activity limitations
  \item \texttt{get\_social\_history} — smoking, alcohol, living situation, occupation
  \item \texttt{get\_pain\_assessment} — pain location, character, severity scale
\end{itemize}
\textbf{Query:} \texttt{"\{\{symptom keyword or drug name, if applicable\}\}"}

Respond as the patient:
\texttt{"""}

\end{tcolorbox}
\caption{Prompts for the Patient Agent}
\end{figure*}


\begin{figure*}[p]
\centering
\begin{tcolorbox}[
  enhanced,
  colback=white, colframe=black!60, arc=2pt, boxrule=0.7pt,
  title={\small\bfseries Checker Planner (Gate 1 — Plan Validation)},
  coltitle=white, colbacktitle=black!78, fonttitle=\small\bfseries,
  top=6pt, bottom=6pt, left=8pt, right=8pt,
]
\small

\textbf{[System Prompt]}=\texttt{"""}

You are a clinical planning validator. Given a user question, task type,
tool source, and an action plan, check whether the plan is correct.

\medskip
Return JSON: \texttt{\{"valid": <bool>, "reason": "<brief explanation>"\}}

\medskip
\textbf{Validation rules:}
\begin{enumerate}[leftmargin=1.8em, topsep=2pt, itemsep=2pt, parsep=0pt]
  \item \textbf{Information Lookup:} exactly 1 non-\texttt{prepare\_to\_answer} tool.\\
    \texttt{tool\_source=ehr} → EHR tool;\quad \texttt{tool\_source=patient} → \texttt{patient.xxx} tool.
  \item \textbf{Data Gathering:} 2–4 non-\texttt{prepare\_to\_answer} tools.\\
    \texttt{tool\_source=mixed}: may combine EHR and \texttt{patient.xxx} tools.
  \item \textbf{Clinical Reasoning:} $\geq$1 non-\texttt{prepare\_to\_answer} tool.\\
    Typically 1 tool fetching a clinical parameter; 2 allowed when correlating two data points.
  \item \textbf{Write/Update:} exactly 1 write tool
    (\texttt{MedicationRequest.create}, \texttt{ServiceRequest.create}, or \texttt{Flag.create})
    $+$ \texttt{prepare\_to\_answer}. No read/search tools allowed.
  \item All tools must exist in the \textbf{Available Tools} list.
  \item \texttt{subject\_id} must be present in EHR tool arguments when the patient is known.
  \item Patient tools (\texttt{patient.xxx}): both \texttt{subject\_id} and \texttt{session\_id} required.
  \item Tool arguments must match their schema (no missing required parameters).
  \item The tools chosen must be \textbf{relevant} to the question asked.
  \item \texttt{Action\_List} must end with \texttt{prepare\_to\_answer}.
\end{enumerate}

Output \textbf{only} the JSON, no other text.
\texttt{"""}

\bigskip

\textbf{[User Prompt]}=\texttt{"""}

\textbf{User question:} \texttt{"\{\{user\_question\}\}"}

\textbf{Task type:} \texttt{\{\{Information Lookup / Data Gathering / Clinical Reasoning / Write/Update\}\}}

\textbf{Tool source:} \texttt{\{\{ehr / patient / mixed / write\}\}}

\textbf{Plan:} \texttt{\{\{Action\_List from Planner Agent\}\}}

\textbf{Available tools:} \texttt{\{\{tool name list\}\}}

Is this plan correct and appropriate? Output JSON.
\texttt{"""}

\bigskip\hrule\vspace{4pt}

\textbf{Two-stage validation:} Rule-based structural checks run first
(tool count per task type, \texttt{subject\_id} presence, tool name membership).
Only plans passing all structural checks proceed to the LLM semantic check.
On \texttt{valid=false}, the Planner Agent is re-invoked with the rejection reason
injected as a warning (up to 3 retries per turn).

\end{tcolorbox}
\caption{Prompts for the Planner Checker Agent}
\end{figure*}


\begin{figure*}[p]
\centering
\begin{tcolorbox}[
  enhanced,
  colback=white, colframe=black!60, arc=2pt, boxrule=0.7pt,
  title={\small\bfseries Clinical Checker (Gate 3 — Answer Validation)},
  coltitle=white, colbacktitle=black!78, fonttitle=\small\bfseries,
  top=6pt, bottom=6pt, left=8pt, right=8pt,
]
\small

\textbf{[System Prompt]}=\texttt{"""}

You are a clinical quality reviewer. Given a clinician's question,
the tool observations (real EHR data), and the AI assistant's answer, check for:

\begin{enumerate}[leftmargin=1.8em, topsep=2pt, itemsep=2pt, parsep=0pt]
  \item \textbf{Hallucination:} Does the answer cite values \textbf{not} present in the observations?
  \item \textbf{Contradiction:} Does the answer contradict values in the observations?
  \item \textbf{Safety:} Does the answer make any obviously dangerous clinical recommendations?
  \item \textbf{Completeness:} Does the answer address the actual question asked?
\end{enumerate}

\medskip
Return JSON:
\begin{itemize}[leftmargin=1.8em, topsep=2pt, itemsep=1pt, parsep=0pt]
  \item \texttt{valid}: boolean
  \item \texttt{hallucination}: boolean
  \item \texttt{contradiction}: boolean
  \item \texttt{safety\_issue}: boolean
  \item \texttt{incomplete}: boolean
  \item \texttt{issues}: list of issue strings
  \item \texttt{score}: integer 0–10
\end{itemize}

Be \textbf{lenient} — minor omissions are acceptable.
Flag \textbf{only} clear errors. A \texttt{safety\_issue=true} unconditionally sets \texttt{valid=false}.
\texttt{"""}

\bigskip

\textbf{[User Prompt]}=\texttt{"""}

\textbf{Clinical Task:} \texttt{\{\{task\_type / scenario\}\}}

\textbf{Question asked:} \texttt{"\{\{user\_question\}\}"}

\textbf{[EHR Data Retrieved]} \textit{(ground truth — truncated to 800 chars per tool call)}\\
\texttt{[\{\{tool\_name\}\}]: \{\{FHIR R4 Bundle JSON\}\}}

\textbf{AI Assistant's answer:}\\
\texttt{\{\{generated\_answer\}\}}

Validate this answer. Output JSON.
\texttt{"""}

\bigskip\hrule\vspace{4pt}

\textbf{Retry logic:} on \texttt{valid=false}, the failed answer is discarded and the Answer Agent
is re-invoked (up to 3 retries). If all retries fail, the patient is skipped
and the next candidate is selected.

\end{tcolorbox}
\caption{Prompts for the Clinical Checker Agent}
\end{figure*}


\begin{figure*}[p]
\centering
\begin{tcolorbox}[
  enhanced,
  colback=white, colframe=black!60, arc=2pt, boxrule=0.7pt,
  title={\small\bfseries Answer Agent},
  coltitle=white, colbacktitle=black!78, fonttitle=\small\bfseries,
  top=6pt, bottom=6pt, left=8pt, right=8pt,
]
\small

\textbf{[System Prompt — Information Lookup]}=\texttt{"""}

You are an AI clinical assistant reporting EHR data to a clinician.

\textbf{Format rules:}
\begin{itemize}[leftmargin=1.8em, topsep=2pt, itemsep=1pt, parsep=0pt]
  \item Report \textbf{only} the directly retrieved value(s).
  \item Format each item as: \texttt{[Item]: [Value] [Unit] ($\uparrow$/$\downarrow$/normal)} — one line per item.
  \item No introductory sentences, no closing remarks, no clinical commentary
    unless the question explicitly asks for interpretation.
  \item If data is missing: \texttt{[Item]: not found in EHR}
  \item Maximum 2 lines total.
\end{itemize}
\texttt{"""}

\medskip\hrule\medskip

\textbf{[System Prompt — Data Gathering]}=\texttt{"""}

You are an AI clinical assistant synthesizing multi-source EHR findings.

\textbf{Format rules:}
\begin{itemize}[leftmargin=1.8em, topsep=2pt, itemsep=1pt, parsep=0pt]
  \item Bullet list, maximum 3 bullets: \texttt{• [Finding]: [clinical implication]}
  \item No introductory or closing sentences.
  \item If a tool returned no data: \texttt{• [item]: not available in EHR}
\end{itemize}

\textbf{Clinical scoring} \textit{(when retrieved data contains scoring components)}:
\begin{itemize}[leftmargin=1.8em, topsep=2pt, itemsep=0pt, parsep=0pt]
  \item Compute score inline using the exact value-to-subscore lookup tables
    (SOFA, SIRS, MELD, Child-Pugh, CURB-65, CHA\textsubscript{2}DS\textsubscript{2}-VASc, HAS-BLED, Ranson,
    Cockcroft-Gault, Wells PE).
  \item Format: component = value → X pts; sum total → category.
  \item \textbf{Critical:} do \textbf{not} default to 2\,pts for "abnormal" — look up the exact range.
  \item List missing components as assumed 0 (note in answer).
\end{itemize}
\texttt{"""}

\medskip\hrule\medskip

\textbf{[System Prompt — Clinical Reasoning]}=\texttt{"""}

You are an AI clinical assistant combining a retrieved patient value with clinical knowledge.

\textbf{Format rules:}
\begin{itemize}[leftmargin=1.8em, topsep=2pt, itemsep=1pt, parsep=0pt]
  \item Respond in \textbf{exactly 2 sentences} — no more.
  \item Sentence 1: state the retrieved patient value with units and whether it is normal/abnormal.
  \item Sentence 2: give \textbf{one} specific, actionable clinical recommendation based on that value.
  \item Do \textbf{not} give generic advice. Do \textbf{not} add a third sentence.
\end{itemize}
\texttt{"""}

\bigskip

\textbf{[User Prompt]}=\texttt{"""} \textit{(shared across all task types)}

\textbf{Task type:} \texttt{\{\{Information Lookup / Data Gathering / Clinical Reasoning\}\}}

\textbf{[Conversation History]}\\
\texttt{\{\{Prior physician queries and assistant responses\}\}}

\textbf{[Tool Observations]} \textit{(real MIMIC-IV data returned by executed tool calls)}\\
\texttt{[\{\{tool\_name\}\}]: \{\{FHIR R4 Bundle JSON — ground truth EHR values\}\}}

\textbf{Physician question:} \texttt{"\{\{user\_question\}\}"}

Generate the clinical response:
\texttt{"""}

\end{tcolorbox}
\caption{Prompts for the Answer Agent}
\end{figure*}


\begin{figure*}[p]
\centering
\begin{tcolorbox}[
  enhanced,
  colback=white, colframe=black!60, arc=2pt, boxrule=0.7pt,
  title={\small\bfseries Rubric Generator Agent},
  coltitle=white, colbacktitle=black!78, fonttitle=\small\bfseries,
  top=6pt, bottom=6pt, left=8pt, right=8pt,
]
\small

\textbf{[System Prompt — General Turns (IL / DG / CR)]}=\texttt{"""}

You are a clinical benchmark rubric designer for an EHR-based QA evaluation.
Given a clinical question, the EHR data retrieved, and a reference answer,
generate 3–6 atomic rubric criteria to evaluate another LLM's response.

\medskip
\textbf{Design rules:}
\begin{enumerate}[leftmargin=1.8em, topsep=2pt, itemsep=2pt, parsep=0pt]
  \item Each item describes an \textbf{outcome or clinical goal} — never a tool call, API name, or process step.
  \item \textbf{Ground items in actual EHR values.}\\
    Write: \textit{``The answer correctly cites creatinine as 0.9\,mg/dL''} — not \textit{``mentions the creatinine value''}
  \item Each item must be independently evaluable as \textbf{YES} or \textbf{NO}.
  \item Include $\geq$1 reasoning or recommendation item (not just fact retrieval).
  \item For safety-critical decisions, include one item checking a dangerous recommendation is \textbf{absent}.
  \item Do \textbf{not} mention tool names, function names, or system internals.
  \item \textbf{Clinical accuracy:} verify the reference answer's conclusions before echoing them.
    If a claim is debatable, write the rubric to check the \textit{reasoning process}, not the specific conclusion.
  \item \textbf{Mixed/patient turns:} cover \textbf{both} dimensions —
    (a) EHR data cited and interpreted correctly;
    (b) patient-reported symptoms/adherence quoted and clinically interpreted.
\end{enumerate}

\medskip
\textbf{Item count by task type:}
\begin{itemize}[leftmargin=1.8em, topsep=2pt, itemsep=1pt, parsep=0pt]
  \item \textbf{IL:} 3 items — value cited, value interpreted, conclusion stated
  \item \textbf{DG:} 4–5 items — each value cited, relationship stated, conclusion
  \item \textbf{CR:} 5–6 items — value cited, threshold applied, reasoning chain, recommendation, safety check
  \item \textbf{Mixed/Patient:} 5–6 items — 2–3 on EHR findings, 2–3 on patient-reported findings
\end{itemize}

Output \textbf{only} a valid JSON array of strings. No prose, no markdown.\\
Example: \texttt{["The answer cites creatinine as 0.9 mg/dL", "The answer concludes no metformin adjustment is needed"]}
\texttt{"""}

\medskip\hrule\medskip

\textbf{[System Prompt — Write/Update Turns (T3 Action)]}=\texttt{"""}

You design rubric items evaluating a model's FHIR write tool call
(\texttt{MedicationRequest.create}, \texttt{ServiceRequest.create}, or \texttt{Flag.create}).

\textbf{Design rules:}
\begin{enumerate}[leftmargin=1.8em, topsep=2pt, itemsep=1pt, parsep=0pt]
  \item Each item names a \textbf{specific field} of \texttt{tool\_call.arguments}
    (e.g.\ \texttt{medication}, \texttt{dose}, \texttt{route}, \texttt{frequency}, \texttt{indication}).
  \item Each item has a clear \textbf{PASS / FAIL} criterion checkable from the field value.
  \item Explicitly allow clinically equivalent values (drug synonyms, dose ranges, frequency synonyms).
  \item Include exactly one \textbf{negative safety item} that FAILS when a dangerous value is present
    (e.g.\ dose $\geq$ contraindicated threshold, wrong drug class, unjustified \texttt{stat} priority).
  \item Do \textbf{not} write items about clinical reasoning or prose justification.
\end{enumerate}
\texttt{"""}

\bigskip

\textbf{[User Prompt]}=\texttt{"""} \textit{(shared across all turn types)}

\textbf{Clinical question:} \texttt{"\{\{user\_question\}\}"}

\textbf{[EHR Data Retrieved]}\\
\texttt{[\{\{tool\_name\}\}]: \{\{actual FHIR R4 Bundle values used in the gold answer\}\}}

\textbf{[Reference Answer]}\\
\texttt{\{\{gold\_answer generated by Answer Agent\}\}}

Generate the rubric criteria:
\texttt{"""}

\end{tcolorbox}
\caption{Prompts for the Rubric Generation Agent}
\end{figure*}

\setcounter{table}{0}
\setcounter{figure}{0}
\section{Full Experiment Results}
\label{sec:full-results}

\subsection{Performance by Clinical Scenario}

Table~\ref{tab:scenario_breakdown} reports rubric scores broken down
by the four clinical scenarios.
Discharge Planning is consistently the hardest scenario across all
models (column average 57.3\%), likely because it requires integrating
longitudinal EHR context, patient preferences, and multi-step care
coordination rather than a single lookup.
Treatment Response and Medication Safety are comparatively easier
(68.1\% and 67.1\%), and show the largest spread between strong and
weak models.
Notably, Qwen3.5-35B-A3B achieves 74.9\% on Medication Safety, on
par with GLM-5 and well above its overall average (66.0\%), suggesting
that some open-weight models have uneven scenario-level strengths.

\begin{table*}[t]
\centering\footnotesize
\setlength{\tabcolsep}{4pt}
\renewcommand{\arraystretch}{1.08}
\resizebox{\textwidth}{!}{%
\begin{tabular}{@{}l cccc c | cccc c@{}}
\toprule
\multirow{2}{*}{\textbf{Model}}
  & \multicolumn{5}{c|}{\textbf{English}}
  & \multicolumn{5}{c}{\textbf{Chinese}} \\
\cmidrule(lr){2-6}\cmidrule(l){7-11}
  & \textbf{Diag.} & \textbf{Dischg.} & \textbf{Med.} & \textbf{Treat.} & \textbf{Avg}
  & \textbf{Diag.} & \textbf{Dischg.} & \textbf{Med.} & \textbf{Treat.} & \textbf{Avg} \\
\midrule
\multicolumn{11}{@{}l}{\textbf{Closed-Source}} \\
\midrule
Claude-Opus-4.7      & \color{blue}{67.7} & 60.8 & 71.1 & 73.4 & 68.3 & \color{red}{70.1} & 60.2 & 71.8 & \color{red}{77.5} & 69.9 \\
GPT-5.4-high         & 61.8 & 59.9 & 68.1 & 66.6 & 64.1 & 66.5 & 62.3 & 67.4 & 67.4 & 65.9 \\
Gemini-3.1-Pro       & 58.9 & 57.5 & 63.7 & 68.3 & 62.1 & 66.0 & \color{blue}{64.1} & 69.4 & 72.5 & 68.0 \\
Seed-1.8             & \color{red}{67.8} & 60.1 & 68.2 & 73.4 & 67.4 & \color{blue}{69.5} & 63.4 & 68.1 & 73.7 & 68.7 \\
GPT-5.4-mini         & 56.3 & 52.0 & 60.3 & 62.4 & 57.8 & 61.6 & 56.1 & 63.6 & 64.8 & 61.5 \\
\midrule
\multicolumn{11}{@{}l}{\textbf{Open-Weight}} \\
\midrule
GLM-5                & 67.6 & \color{blue}{61.6} & \color{blue}{74.6} & 73.7 & 69.4 & 68.9 & \color{red}{64.7} & \color{red}{75.1} & \color{blue}{77.4} & 71.5 \\
Kimi-K2.6            & 64.7 & \color{red}{61.8} & 71.3 & \color{blue}{74.3} & 68.0 & 65.5 & 63.2 & \color{blue}{74.9} & 75.9 & 69.9 \\
DS-V4-Pro            & 65.2 & 59.3 & 72.6 & \color{red}{75.1} & 68.1 & 62.4 & 60.3 & 66.3 & 70.2 & 64.8 \\
DS-V4-Flash          & 62.1 & 54.0 & 70.3 & 70.7 & 64.3 & 54.6 & 58.7 & 63.5 & 65.5 & 60.6 \\
MiniMax-M2.7         & 64.2 & 55.2 & 66.5 & 66.9 & 63.2 & 67.2 & 58.9 & 69.5 & 71.6 & 66.8 \\
Qwen3.5-35B-A3B      & 61.7 & 58.4 & \color{red}{75.1} & 70.0 & 66.3 & 62.2 & 54.5 & 74.7 & 71.4 & 65.8 \\
Qwen3.5-27B          & 56.7 & 50.9 & 66.0 & 59.7 & 58.3 & 61.7 & 49.8 & 68.2 & 65.9 & 61.4 \\
Qwen3.5-9B           & 55.9 & 52.2 & 56.6 & 56.1 & 55.2 & 58.5 & 53.7 & 60.2 & 59.6 & 58.0 \\
Qwen3.5-4B           & 44.3 & 47.5 & 52.1 & 50.9 & 48.7 & 44.8 & 43.2 & 49.1 & 52.5 & 47.4 \\
\bottomrule
\end{tabular}}
\caption{Rubric score (\%) per clinical scenario, English and Chinese separately.
Diag.=Diagnostic Workup, Dischg.=Discharge Planning, Med.=Medication Safety, Treat.=Treatment Response.
Numbers in {\color{red}red} and {\color{blue}blue} are the best and second-best per column within each language group.}
\label{tab:scenario_breakdown}
\end{table*}

\subsection{Performance by Data Richness}

Table~\ref{tab:difficulty_breakdown} reports rubric scores across
three data richness tiers (High / Medium / Low).
Tier gaps are small (typically 2--3 pp), and, counter-intuitively,
High-richness sessions score slightly above Low-richness ones
(64.5\% vs.\ 63.5\%), suggesting denser records offer disambiguation
cues that offset added complexity.
Weaker models (Qwen3.5-9B, Qwen3.5-4B) trend downward with declining
richness while stronger models do not, implying richness matters more
when capacity is limited.
However, these differences are descriptive rather than causal, since
tier assignment changes evidence availability, case composition, and
required retrieval patterns.

\begin{table*}[t]
\centering
\setlength{\tabcolsep}{5pt}
\renewcommand{\arraystretch}{1.08}
\begin{tabular}{@{}l ccc c | ccc c@{}}
\toprule
\multirow{2}{*}{\textbf{Model}}
  & \multicolumn{4}{c|}{\textbf{English}}
  & \multicolumn{4}{c}{\textbf{Chinese}} \\
\cmidrule(lr){2-5}\cmidrule(l){6-9}
  & \textbf{High} & \textbf{Med.} & \textbf{Low} & \textbf{Avg}
  & \textbf{High} & \textbf{Med.} & \textbf{Low} & \textbf{Avg} \\
\midrule
\multicolumn{9}{@{}l}{\textbf{Closed-Source}} \\
\midrule
Claude-Opus-4.7      & 69.3 & 67.7 & 67.8 & 68.3 & \color{blue}{71.2} & 67.5 & \color{blue}{70.9} & 69.9 \\
GPT-5.4-high         & 63.7 & 64.9 & 63.6 & 64.1 & 66.4 & 64.9 & 66.4 & 65.9 \\
Gemini-3.1-Pro       & 62.3 & 61.2 & 62.8 & 62.1 & 67.1 & 68.3 & 68.6 & 68.0 \\
Seed-1.8             & 66.7 & 67.6 & 68.0 & 67.4 & 68.1 & 68.0 & 69.9 & 68.7 \\
GPT-5.4-mini         & 59.8 & 57.1 & 56.4 & 57.8 & 64.8 & 59.2 & 60.7 & 61.5 \\
\midrule
\multicolumn{9}{@{}l}{\textbf{Open-Weight}} \\
\midrule
GLM-5                & \color{red}{71.0} & \color{red}{69.1} & \color{blue}{68.2} & 69.4 & \color{red}{72.7} & \color{red}{70.3} & \color{red}{71.6} & 71.5 \\
Kimi-K2.6            & \color{blue}{69.4} & \color{blue}{68.1} & 66.5 & 68.0 & 71.0 & \color{blue}{68.9} & 69.7 & 69.9 \\
DS-V4-Pro            & 68.3 & 67.2 & \color{red}{68.7} & 68.1 & 67.0 & 61.5 & 65.9 & 64.8 \\
DS-V4-Flash          & 65.7 & 63.1 & 64.0 & 64.3 & 62.2 & 62.0 & 57.6 & 60.6 \\
MiniMax-M2.7         & 62.5 & 62.4 & 64.7 & 63.2 & 67.5 & 66.8 & 66.2 & 66.8 \\
Qwen3.5-35B-A3B      & 66.4 & 65.7 & 66.9 & 66.3 & 65.1 & 66.2 & 65.8 & 65.8 \\
Qwen3.5-27B          & 60.7 & 56.4 & 57.8 & 58.3 & 62.2 & 60.7 & 61.3 & 61.4 \\
Qwen3.5-9B           & 57.1 & 53.5 & 55.0 & 55.2 & 59.4 & 56.8 & 57.8 & 58.0 \\
Qwen3.5-4B           & 49.9 & 47.5 & 48.8 & 48.7 & 48.2 & 47.2 & 46.8 & 47.4 \\
\bottomrule
\end{tabular}
\caption{Rubric score (\%) per data richness tier, English and Chinese separately.
High / Medium / Low correspond to EHR record completeness levels 1--3.
Numbers in {\color{red}red} and {\color{blue}blue} are the best and second-best per column within each language group.}
\label{tab:difficulty_breakdown}
\end{table*}

\subsection{Pass Rate Curves across Thresholds}

\begin{figure*}[!htbp]
  \centering
  \includegraphics[width=\linewidth]{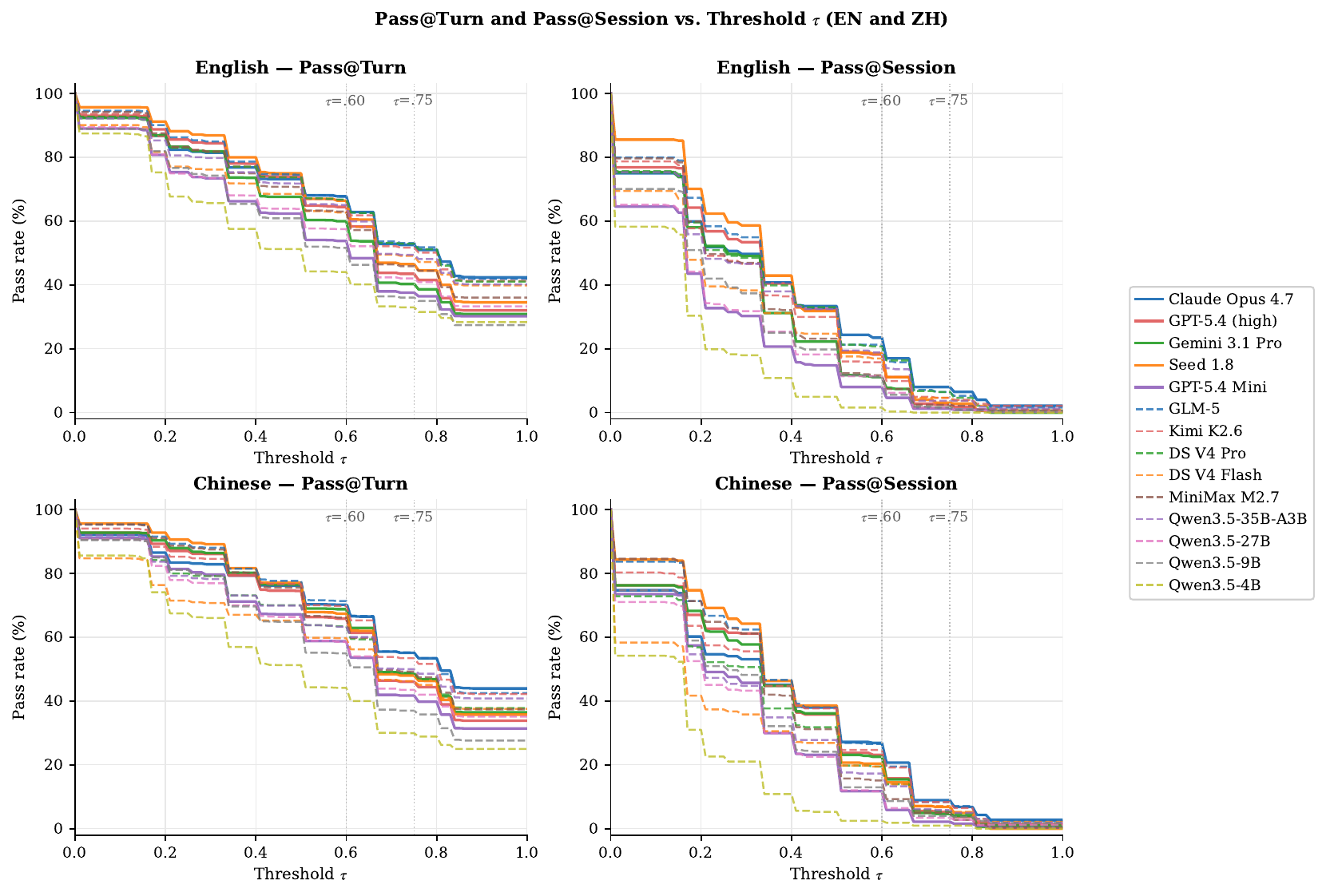}
  \caption{Pass@Turn (left column) and Pass@Session (right column) as a
    function of threshold $\tau$ for English (top) and Chinese (bottom).
    Solid lines = closed-source models; dashed lines = open-weight models.
    Vertical dotted lines mark $\tau{=}0.60$ and $\tau{=}0.75$.}
  \label{fig:pass-threshold}
\end{figure*}

Figure~\ref{fig:pass-threshold} plots Pass@Turn and Pass@Session
against threshold $\tau \in [0,1]$ for all fourteen models,
separately for English and Chinese.
Three patterns hold across both languages.
First, Pass@Session decays far faster than Pass@Turn: at $\tau{=}0.60$,
Claude-Opus-4.7 reaches 67.7\% Pass@Turn (EN) but only 23.5\%
Pass@Session, and by $\tau{=}0.75$ most models drop below 10\%
Pass@Session even with Pass@Turn above 40\%.
Second, the spread between strong and weak models widens at the
session level: Qwen3.5-4B records 1.5\% Pass@Session at $\tau{=}0.60$
(EN), approximately 20 pp below Claude and GLM-5.
Third, Chinese scores stay slightly higher than English at every
threshold (e.g., Claude 23.5\%$\to$26.9\%, GLM-5 21.3\%$\to$26.5\% at
$\tau{=}0.60$).

\begin{figure*}
    \centering
    \includegraphics[width=\linewidth]{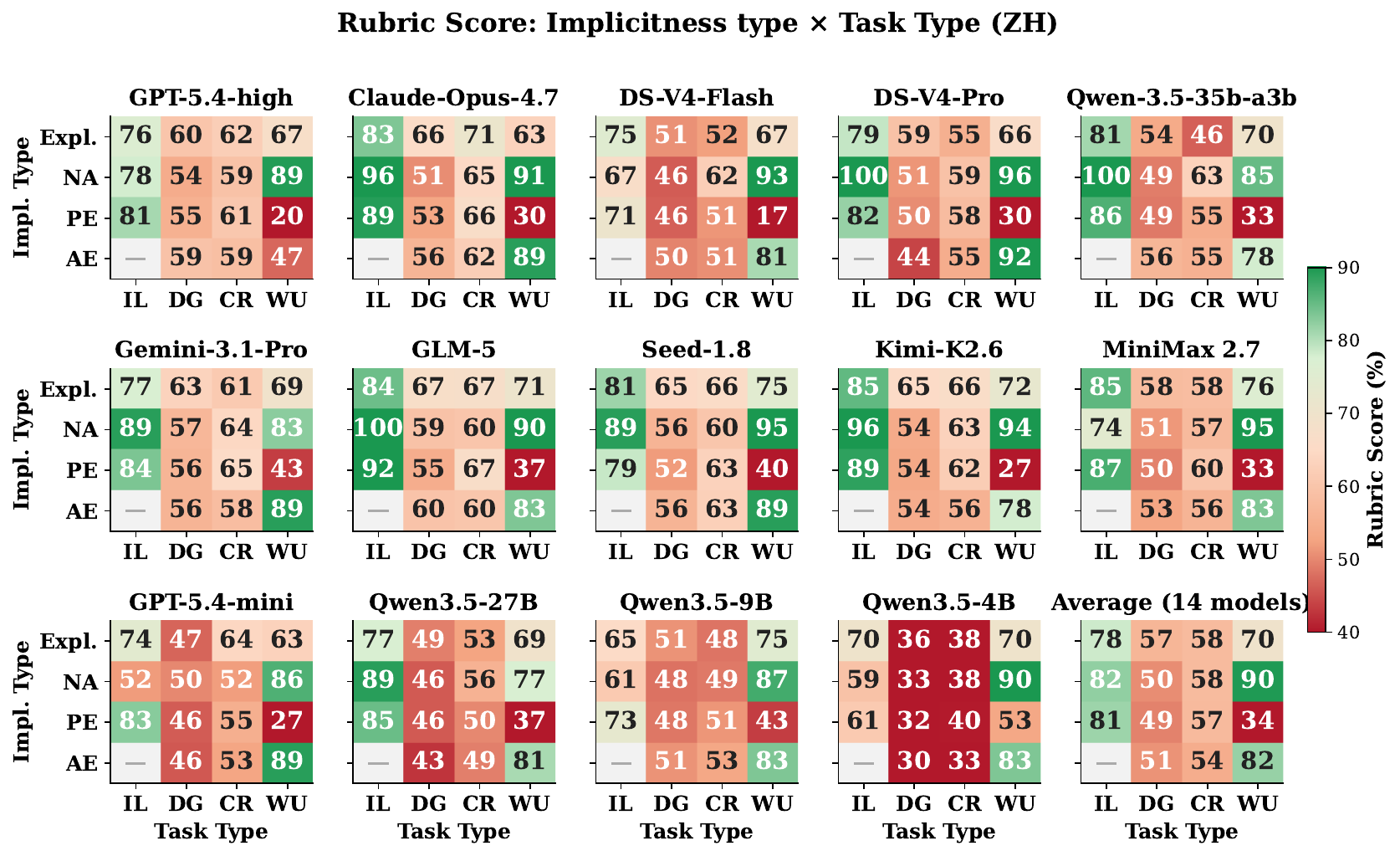}
    \caption{Rubric score (\%) by implicitness type × task type (ZH), for 14 models and their average.}
    \label{fig:heatzh}
\end{figure*}

\setcounter{table}{0}
\setcounter{figure}{0}

\end{document}